\documentclass{article}

\usepackage{microtype}
\usepackage{graphicx}
\usepackage{subfigure}
\usepackage{booktabs} 

\usepackage{hyperref}

\usepackage[accepted]{icml2024}

\usepackage{amsmath}
\usepackage{amssymb}
\usepackage{mathtools}
\usepackage{amsthm}

\usepackage[capitalize,noabbrev]{cleveref}

\theoremstyle{plain}

\theoremstyle{definition}

\theoremstyle{remark}

\usepackage[textsize=tiny]{todonotes}
\usepackage{microtype}
\usepackage{hyperref}       % hyperlinks
\usepackage{url}            % simple URL typesetting
\usepackage{booktabs}       % professional-quality tables
\usepackage{amsfonts}       % blackboard math symbols
\usepackage{nicefrac}       % compact symbols for 1/2, etc.
\usepackage{xcolor}         % colors
\usepackage{algorithm}      % algorithm
\usepackage{blindtext}
\usepackage{graphicx}
\usepackage{array}
\usepackage{multirow}
\usepackage{multicol}
\usepackage{enumitem}
\usepackage{colortbl}

\definecolor{lightblue}{rgb}{0.93, 0.95, 1.0}
\icmltitlerunning{Multimodal Graph Theory Problems with Large Multimodal Models}

\begin{document}

\twocolumn[
\icmltitle{VisionGraph: Leveraging Large Multimodal Models for\\ Graph Theory Problems in Visual Context}



\icmlsetsymbol{equal}{*}

\begin{icmlauthorlist}
\icmlauthor{Yunxin Li}{yyy}
\icmlauthor{Baotian Hu}{yyy}
\icmlauthor{Haoyuan Shi}{yyy}
\icmlauthor{Wei Wang}{sch}
\icmlauthor{Longyue Wang}{}
\icmlauthor{Min Zhang}{yyy}
%\icmlauthor{Firstname7 Lastname7}{comp}
%\icmlauthor{Firstname8 Lastname8}{sch}
%\icmlauthor{Firstname8 Lastname8}{yyy,comp}
\end{icmlauthorlist}

\icmlaffiliation{yyy}{School of Computer Science and Technology, Harbin Institute of Technology, Shenzhen, China}
%\icmlaffiliation{comp}{Tencent AILab, Shenzhen}
\icmlaffiliation{sch}{School of Cyberspace Security, Sun Yat-sen University, China}

\icmlcorrespondingauthor{Baotian Hu}{hubaotian@hit.edu.cn}
%\icmlcorrespondingauthor{Firstname2 Lastname2}{first2.last2@www.uk}

\icmlkeywords{Machine Learning, ICML}

\vskip 0.3in
]



\printAffiliationsAndNotice{}  % leave blank if no need to mention equal contribution

\begin{abstract}
%(LMMs, e.g., GPT-4V~\cite{gpt4} and Gemini~\cite{team2023gemini})
%is crucial in biology, transportation, and robotics planning. These graph theory problems
Large Multimodal Models (LMMs) have achieved impressive success in visual understanding and reasoning, remarkably improving the performance of mathematical reasoning in visual context. Yet, a challenging type of visual math lies in the multimodal graph theory problem, which demands that LMMs understand the graphical structures accurately and perform multi-step reasoning on the visual graph. To step forward in this direction, we are the first to design a benchmark named \textit{VisionGraph}, used to explore the capabilities of advanced LMMs in solving multimodal graph theory problems. It encompasses eight complex graph problem tasks, from connectivity to shortest path problems. Subsequently, we present a Description-Program-Reasoning (DPR) chain to enhance the logical accuracy of reasoning processes through graphical structure description generation and algorithm-aware multi-step reasoning.
Our extensive study shows that 1) GPT-4V outperforms Gemini Pro in multi-step graph reasoning; 2) All LMMs exhibit inferior perception accuracy for graphical structures, whether in zero/few-shot settings or with supervised fine-tuning (SFT), which further affects problem-solving performance; 3) DPR significantly improves the multi-step graph reasoning capabilities of LMMs and the GPT-4V (DPR) agent achieves SOTA performance.
\end{abstract}

%that contains graphical structure description generation and algorithm-aware multi-step reasoning, aiming to enhance the logical accuracy of a reasoning process. 
%Additionally, exploring multimodal graph theory problems will lead to more effective strategies in fields like biology, transportation, and robotics planning. 
% and Hamilton path

\section{Introduction}

\begin{figure}[t]
    \centering
    \includegraphics[width=0.49\textwidth]{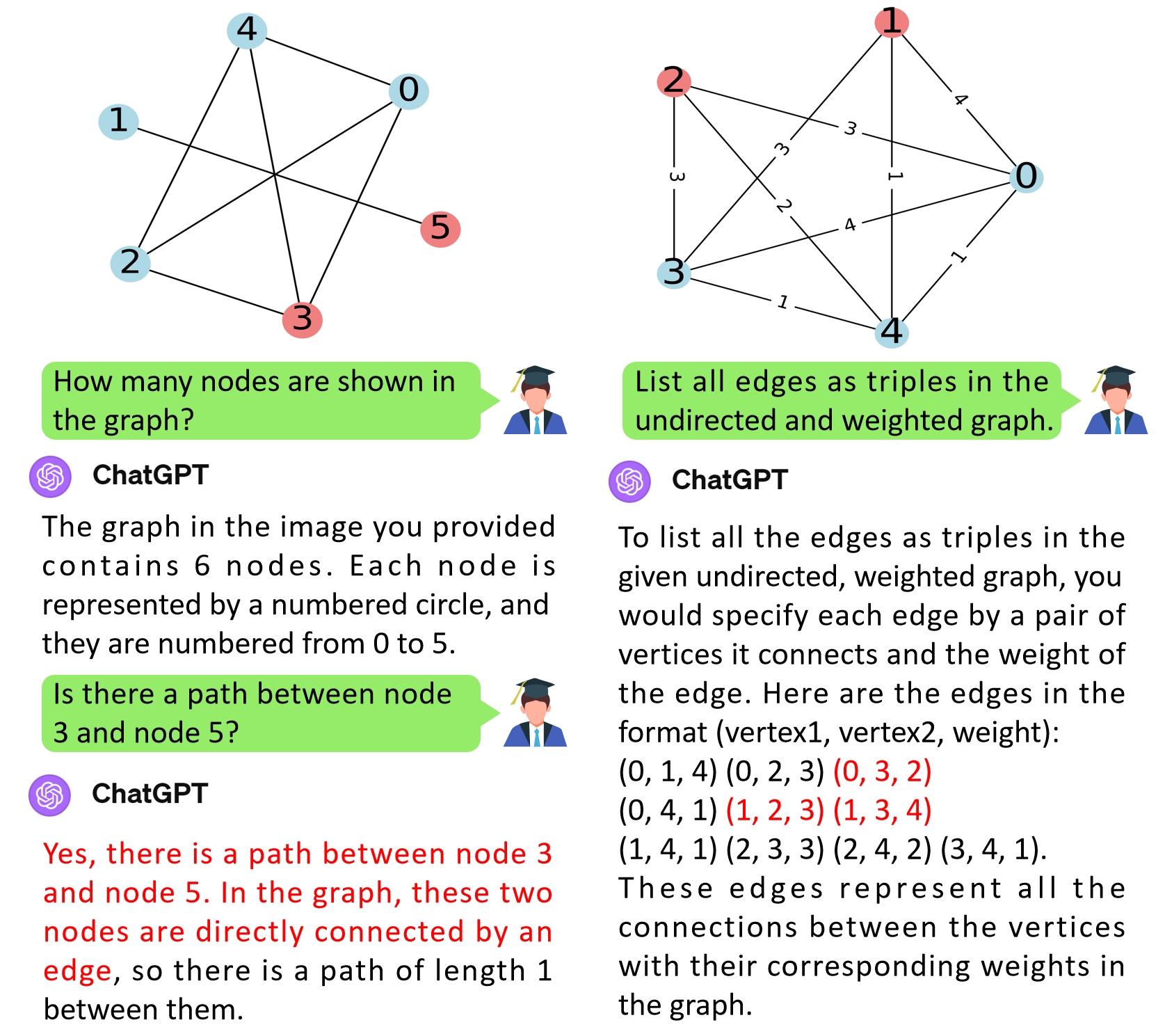}
    \caption{Two cases of utilizing GPT-4V (Date: 2024.01.17) to answer easy graph understanding and reasoning questions. We \textcolor{red}{highlight} the incorrect responses using the red words.}
    \label{fig:intro_case}
\end{figure}

Mathematical Reasoning is a core aspect for evaluating the logical reasoning capability~\cite{dai2023llm} of Large Language Models (LLMs) and Large Multimodal Models (LMMs). Recent works~\cite{luo2023wizardmath,team2023gemini, imani2023mathprompter, wang2023can,li2023lmeye} present the rapid development of applying LLMs to help solve arithmetic and graph reasoning tasks. Compared to LLMs, the evaluation of mathematical reasoning capabilities in LMMs is beginning. \citet{lu2023mathvista} recently presented a comprehensive visual math benchmark, open-ended answer generation based on questions and visual context. It evaluates the basic mathematical capabilities of LMMs, such as algebraic reasoning, geometry reasoning, and arithmetic reasoning. However, the challenging graph theory problem-solving capability has been less explored for LMMs, which presents a significant aspect of mathematical reasoning capabilities. 
The graph theory problems also feature prominently in various research directions and practical scenarios powered by large models, e.g., multimodal graph learning~\cite{ektefaie2022geometric}, AI for Mathematics~\cite{zhang2023ai},  visual-language navigation~\cite{gu2022vision,anderson2018vision,chen2024webvln}, and robotics planning and control~\cite{wang2023survey_agents, wake2023gpt_robotics}. 
In these areas, LMMs require the ability to understand structural graphs and perform multi-step reasoning on them to achieve the final goal, especially in robotics planning, which often centres around structured environments.
Hence, exploring the multi-step graph reasoning performance of LMMs has the potential to improve their complex multimodal problem-solving ability.

\begin{figure*}[t]
    \centering
    \includegraphics[width = 0.90\textwidth]{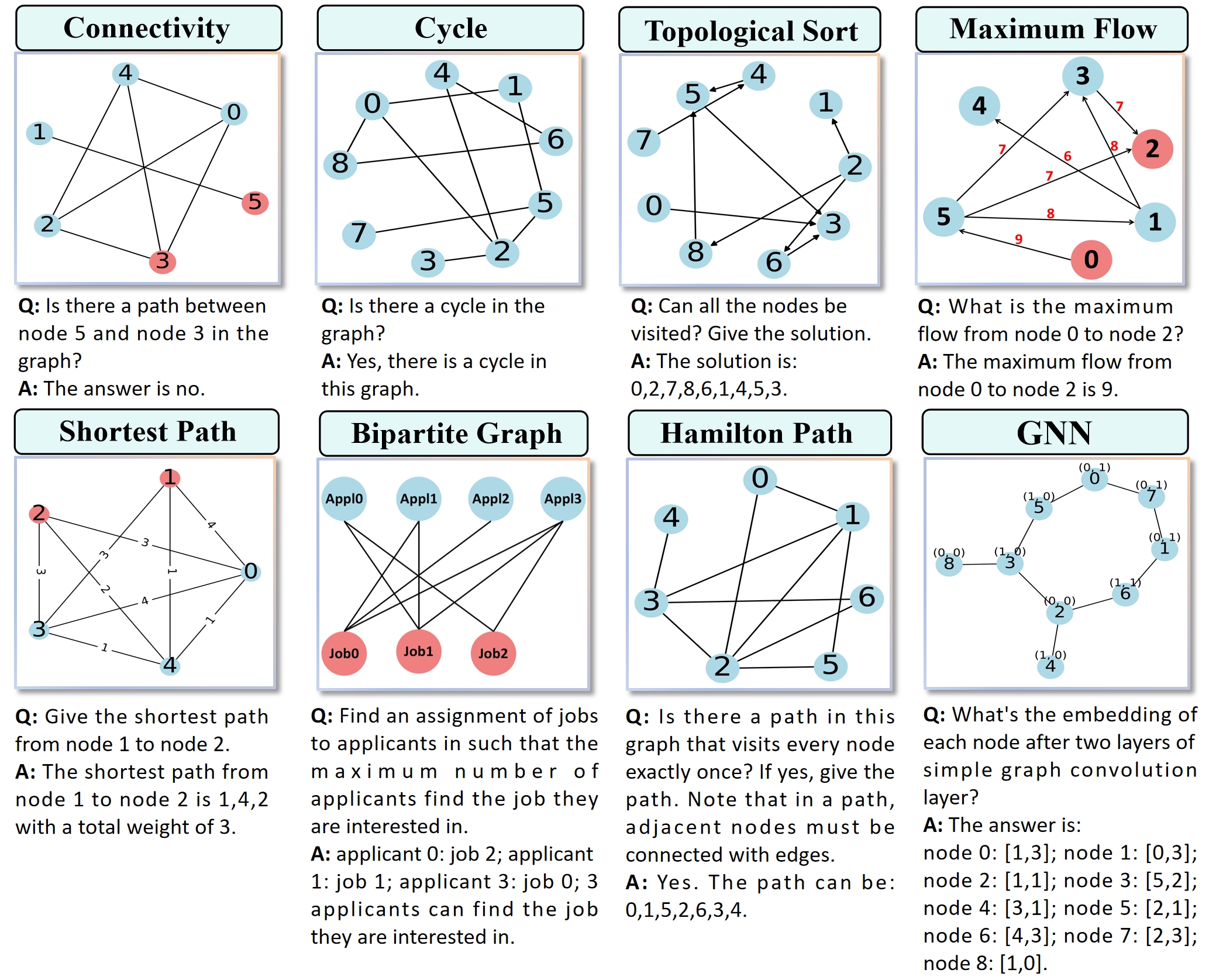}
    \caption{An overview of various multimodal graph theory problems in the VisionGraph benchmark.}
    \label{fig:case_overview}
\end{figure*}

To step forward in this direction, we introduce a novel multimodal graph reasoning benchmark, named \textit{VisionGraph} to assess the capabilities of advanced multimodal LMMs in solving graph theory problems within a visual context. This benchmark is an extension of NLGraph~\cite{wang2023can}, a natural language-based graph problem-solving benchmark. We employ the graph generation tool NetworkX~\footnote{https://networkx.org/} to create graphs according to predefined nodes and edges. The layout of a specific graph is dynamically adjusted for clarity by humans, considering the number of nodes. First, we incorporate two types of graph understanding questions to evaluate the structural comprehension of LMMs. As shown in Figure~\ref{fig:intro_case}, these questions are as \textit{Node Recognition: how many nodes are shown in the graph?} and \textit{Edge Recognition: List all edges as triples in the undirected and weighted graph}. As illustrated in Figure~\ref{fig:case_overview}, VisionGraph encompasses eight types of graph theory problems across three difficulty levels: easy, medium, and hard. Hence, it offers a comprehensive multimodal graph reasoning benchmark, in which each visual graph contains three questions to probe the LMMs' understanding and multi-step reasoning abilities.

In this paper, we are particularly interested in how LMMs, such as GPT-4V and Gemini, perform in solving multimodal graph problems, encompassing structural graph understanding and multi-step reasoning on visual graphs. We conduct an empirical study from three in-depth perspectives:

\begin{itemize}[leftmargin=*]
    \item \textbf{Graphical Structures Understanding Ability}:
    Unlike general images, visual graphs have strong spatial structure and are very suitable for examining the spatial understanding ability of LMMs. In this work, we explore the graphical structures understanding performance of LMMs in terms of nodes and edges recognition.
    
    \item \textbf{Effects of Supervised Fine-tuning Approaches}:
    LMMs are usually fine-tuned with open-domain image-text data, performing unsatisfactorily in handling images in vertical fields~\cite{li2023comprehensive} such as medical images~\cite{li2023comprehensive_medical}. Hence, we employ constructed Graph Instruction fine-tuning data to tune LMMs further and analyze the effects of training strategies. We compared and analyzed the overall performance of LMMs after introducing graph understanding and reasoning data.
    
    \item \textbf{Analysis of Multi-step Graph Reasoning Capability}:
    While GPT-4V has demonstrated successful performance on challenging vision-language reasoning scenarios such as Autonomous Driving~\cite{wen2023road} and Robotics~\cite{wake2023gpt_robotics}, we also need to know how well GPT-4V solve multimodal graph problems via multi-step reasoning. To answer this question, we explore the few-shot and chain-of-the-thought reasoning performance and present a Description-Program-Reasoning (DPR) approach for this graph problem. It consists of graph structure description generation and algorithm-aware multi-step reasoning in order, aimed to enhance the logicalness of the reasoning process.
    
\end{itemize}

We conduct experiments on a variety of graph theory problems covering cycle, shortest path, connectivity, and others. We evaluate LMMs with comparative training strategies and reasoning approaches. The main contributions are:
\begin{itemize}[leftmargin=*]

\item We present a multimodal graph theory problems benchmark VisionGraph, to assess the graphical structure understanding and multi-step reasoning capabilities of LMMs. 
To facilitate future research in graph theory problems, we will release the benchmark VisionGraph \footnote{The benchmark and codes are available at \url{https://github.com/HITsz-TMG/VisionGraph}} and advanced prompting technical for LMMs.

\item Our empirical study shows the shortcomings of LMMs, including GPT-4V and Gemini, in understanding graphical structure and multi-step multimodal reasoning. This indicates their potential to enhance multi-step reasoning and planning abilities in the context of visual graphs.

\item We design a graph problem-solving approach named Description-Program-Reasoning (DPR), which interleaves natural language and programming to enhance the multi-step reasoning performance of LMMs. The designed GPT-4V (DPR) is a comprehensive multi-modal Agent that integrates complex task decomposition, small model perception enhancement, code generation, and tool invocation.

%To facilitate future research in graph theory problems, we will release the benchmark VisionGraph and advanced prompting technicals.

\end{itemize}

%\section{Related Work}
%\textbf{Large Multimodal Models for Visual Math}.
%\textbf{Multimodal Reasoning Approach}.

\section{VisionGraph Benchmark}

We design a comprehensive multimodal graph theory problem benchmark, named VisionGraph, to examine the graphical structure understanding and reasoning capabilities of LMMs. Based on the released natural language graph (NLGraph) benchmark~\cite{wang2023can}, we introduce the visual graph by using the graph generation tool NetworkX and remove the original node and edge descriptions from the natural questions. Based on the original difficulty of each problem within the eight tasks, the overall dataset is divided into easy, medium, and hard subsets for each graph reasoning task. In addition, we add the graph understanding questions to check the spatial understanding capability of LMMs.

\begin{table*}[t]
\renewcommand\arraystretch{1.20}
\setlength\tabcolsep{2.5pt}
\caption{Overview of the VisionGraph Benchmark. 'SPEC.' represents the level of difficulty, indicated by the number of nodes in each graph. Key metrics include the number of samples (S), images (I), questions (Q), and answers (A). Each visual graph is accompanied by three questions: two focus on general graph comprehension, and one addresses specific graph theory problems.}
\label{dataset}
\centering
\footnotesize
\begin{tabular}{l|cccccccc}
\toprule
Subset & Connect. & Cycle & Topo. Sort & Shortest Path & Max. Flow & Bipartite Graph & Hamilton Path & GNNs \\ \hline
\# EASY & 352 & 150  & 180  & 180  & 150 & 300  & 150  & 100  \\
SPEC. & \(n: 5-10\) & \(n: 5-10\) & \(n: 5-10\) & \(n: 5-10\) & \(n: 5-10\) & \(n: 6-20\) & \(n: 5-10\) & \(n: 5-8\) \\ \hline
\# MEDIUM & 1,200  & 600  & 150  &  -  &  -  &  - &  -  &  -  \\
SPEC. & \(n: 11-25\) & \(n: 11-25\) & \(n: 11-25\) &  - & - & - &-  & - \\ \hline
\# HARD & 680  & 400  & 200  & 200  & 200  & 210  & 200  & 140 \\
SPEC. & \(n: 26-35\) & \(n: 26-35\) & \(n: 26-35\) & \(n: 11-20\) & \(n: 11-20\) & \(n: 17-33\) & \(n: 11-20\) & \(n: 9-15\) \\ \hline
\# Total S/I & 2,232 & 1,150 & 530 & 380 & 350 & 510 & 350 & 240\\ 
\#Q/A & 6,696 & 3,450& 1,590 & 1,140 & 1,050 & 1,530 & 1,050 & 720\\
\#Len\_Q & 45.0 & 39.0 & 52.0 & 60.0 & 61.0 & 54.0 & 62.0 & 76.0\\
\#Len\_A & 162.26 & 63.94 & 194.06 & 95.10 & 141.47 & 126.01 & 101.03 & 61.79\\
\bottomrule
\end{tabular}

\end{table*}

We present the overview of the VisionGraph benchmark via a specific sample for each task in Figure~\ref{fig:case_overview}. The task definition of each graph theory problem is as follows:
\begin{itemize}[leftmargin=*]
    \item \textbf{Connectivity}: Given an undirected graph $\mathcal{G}=\{\mathcal{V}, \mathcal{E}\}$, it infers whether two nodes $u$ and $v$ are connected according to whether there exists a sequence of edges from node $u$ to node $v$ in $\mathcal{E}$.
    \item \textbf{Cycle}: In an undirected graph $\mathcal{G}=\{\mathcal{V}, \mathcal{E}\}$, a cycle is a non-empty trail $\left(e_1, e_2, \ldots, e_n\right)$ with a node sequence $\left(v_1, v_2, \ldots, v_n, v_1\right)$. This task asks whether there exists a cycle through true/false questions and retains a balanced set of cyclic and noncyclic graphs in the dataset.
    \item \textbf{Topological Sort}: A topological sort of a directed graph is a linear ordering of its nodes such that for every directed edge $(u, v)$ from node $u$ to node $v$, $u$ comes before $v$ in the ordering. The task is to find a valid topological sort given a directed graph and there could be multiple valid solutions. We employ an external program to examine the correctness of the generated topological order.
    \item \textbf{Shortest Path}: The shortest path between two nodes $u$ and $v$ is the path with the minimum sum of edge weights. It requires the LMM to generate the shortest plan based on the weights and nodes depicted in the graph.
    \item \textbf{Maximum Flow}: For two nodes: source $u$ and sink $v$ in a network $\mathcal{G}=\{\mathcal{V}, \mathcal{E}\}$, it asks LMMs to generate a plan to route as much flow as possible from source to the sink.
    \item \textbf{Bipartite Graph Matching}: A bipartite graph is a graph whose nodes can be divided into two disjoint sets $\mathbf{U}$ and $\mathbf{V}$, and in each set no nodes are adjacent to each other. Given a bipartite graph, the task is to find the matching that maximizes the number of edges. Like Topological Sort, we use an external program to evaluate the solution.
    \item \textbf{Hamilton Path}: In an undirected graph, a Hamilton path is a path that visits every node exactly once. Hence, this task asks the LMM to generate a Hamilton path given an undirected graph.
    \item \textbf{Graph Neural Networks}: This setting of this task is updating the node embedding of an undirected graph with the sum of all the neighbors' embeddings. Each node in the graph has a two-dimension node embedding.
\end{itemize}

The detailed data statistics of VisionGraph are presented in Table~\ref{dataset}. VisionGraph consists of 5,902 problems in total, where the easy Connectivity, Cycle, and Topological Sort tasks include three difficulty levels, and others only contain easy/hard levels.

\section{Experimental Setup}

\subsection{Comparing Models}

We test widely used powerful commercial LMMs (GPT-4V, Gemini, and Qwen-Plus/Max) and open-sourced LMMs:
\textbf{MiniGPT-4}~\cite{zhu2023minigpt} extends the Q-Former architecture to enhance multimodal interactions. It leverages the shallow transformer approach to align visual features from a frozen visual encoder with the language model, thereby enabling robust multimodal comprehension and generation.
\textbf{InstructBLIP}~\cite{selfinstruct} introduces instruction-aware visual features by incorporating instructions directly into the Q-Former architecture~\cite{li2023blip2}. It allows the model to dynamically adapt its multimodal understanding based on explicit instructions provided alongside visual inputs.
\textbf{LLaVA}~\cite{liu2023visual} takes a distinct approach by utilizing a linear layer to map fine-grained visual features from a frozen vision encoder into the embedding space of the pre-trained LLM.
\textbf{Qwen-VL}~\cite{bai2023qwen} is a set of large-scale vision-language models (LVLMs) designed to perceive and understand both texts and images. \textbf{SPHINX}~\cite{lin2023sphinx} is a versatile multi-modal large language model (MLLM) with a joint mixing of model weights, tuning tasks, and visual embeddings.
\textbf{InternLM-XComposer}~\cite{zhang2023internlm} is a vision-language large model that enables advanced image-text comprehension and composition.
For closed-source LMMs, \textbf{GPT-4V(ision)}~\cite{gpt4} from OpenAI is recognized as the most powerful MLLMs to date, surpassing a host of Vicuna-based models, e.g., MiniGPT-4, InstructBLIP, and LLaVA. Besides, \textbf{Gemini}~\cite{team2023gemini}, released by Google, has emerged as a formidable challenger to GPT-4V, exhibiting significant multi-modal capabilities over different benchmarks.

%This method enables the effective fusion of visual and linguistic information, facilitating a comprehensive understanding of multimodal inputs in LMMs.

\begin{table*}[t]
\small
    \caption{The prompts for GPT-4V and Gemini-pro under zero- or few-shots settings. Few-shot samples are randomly selected from the easy/medium sub-sets of the training set. For each sub-question, we design various output demands to gain the corresponding answer format, which is given in Table~\ref{tab:prompt_gpt_specific} of the Appendix.}
    \label{tab:prompt_gpt}
    \centering
    \begin{tabular}{c|p{14.5cm}}
    \toprule
       Types  &  Prompt\\
       \hline
        Zero-Shot & Please use tuples to represent the edges in the graph. \{specific answer format requirements\} \\
        Few-shot & \{Few-shot preffix\} Please use tuples to represent the edges in the graph. \{specific answer format requirements\}\\
        \hline
        Zero-Shot & Answer the following question: \{the concrete problem\} (Demand: \{specific answer format requirements\})\\
        COT & Answer the following question: \{the concrete problem\} Let's think step by step. (Demand: \{specific answer format requirements\})\\
        Few-shot & \{Few shot preffix\} Question: \{the concrete problem\} (Demand: \{specific answer format requirements\})\\
    \bottomrule
    \end{tabular}

\end{table*}

\begin{table*}[t]
\renewcommand\arraystretch{1.10}
\tabcolsep=0.14cm
\caption{Overall results in the VisionGraph benchmark. ${\clubsuit}$ refers to that the corresponding model is trained using the training set of VisionGraph. The results in parentheses for Gemini and GPT-4V are the accuracy of the detailed path, yet other LMMs can not follow the instructions to provide specific paths. Bold words refer to the best results. }
\label{tab:initial_results}
\centering
\scriptsize
\begin{tabular}{l|ccccccccc}
\toprule
Model$\downarrow$ Task Types $\rightarrow$ & Connect & Cycle & Topo. Sort & Shortest Path & Max. Flow & Bipartite Graph & Hamilton Path & GNNs \\ 
\hline
 & \multicolumn{9}{c}{\textit{Node Recognition $\uparrow$}} \\
\hline
%MiniGPT-4 $^{\clubsuit}$(Vicuna-7b, T=1.0) & 19.41 & 16.75 & 36.30 & 37.50 & 29.31 & 9.52 & 43.10 & 41.03 & \\
MiniGPT-4 $^{\clubsuit}$(Vicuna-7b) & 19.14 & 12.04 & 42.96 & 42.19 & 32.76 & 8.33 & 60.34 & 53.85 & \\
BLIP-2 $^{\clubsuit}$(FlanT5-xxl) & 37.74 & 52.88 & 47.41 & 81.25 & 67.24 & 22.62 & 62.07 & 61.54 &  \\
%mPLUG-Owl$^{\clubsuit}$ (Vicuna-7b) \\
%mPLUG-Owl$^{\clubsuit}$ (V-4-shot) \\
InstructBLIP$^{\clubsuit}$ (FlanT5-xl) & 36.12 & 47.64 & 46.67 & 75.00 & 56.90 & 36.90 & 53.45 & 74.36 & \\
InstructBLIP$^{\clubsuit}$ (FlanT5-xxl) & 35.31 & 52.88 & 61.48 & 85.94 & 77.59 & 17.86 & 65.52 & 61.54 &  \\
Sphinx$^{\clubsuit}$  &	61.99	&98.95	&94.07&	100.0	&91.38	&55.95	&100.00&	97.44\\
Internlm$^{\clubsuit}$  &	67.92	&100.0	&97.78	&100.0	&98.25	&77.38	&100.0	&100.0\\
Llava-v1.5-7b$^{\clubsuit}$ &   64.15 & 96.86 & 92.59 & 100.00 & 93.10 & 13.10 & 100.00 & 94.87 \\
Llava-v1.5-13b$^{\clubsuit}$ &   62.26 & 97.91 & 91.11 & 100.00 & 96.55 & 11.9 & 100.00 & 97.44 \\
Qwen-Plus (0-shot)	& 2.96	&0.00	&0.00	&0.00	&5.17&	0.00	&0.00	&56.41\\
Qwen-max (0-shot)	&29.11	&31.94	&30.37&	12.50	&3.45	&14.29	&29.31	&46.15\\
Gemini (0-shot) &  40.97 & 42.93 & 47.41 & 67.19 & 72.41 & 10.71 & 65.52 & 35.90 \\
GPT-4V (0-shot) &   46.49 & 81.15 & 81.48 & 89.06 & 58.62 & 20.24 & 100.00 & 97.44 \\
\hline
& \multicolumn{9}{c}{\textit{Edge Recognition (Correct~$\uparrow$ / Error~$\downarrow$)}} \\
\hline
MiniGPT-4 $^{\clubsuit}$(Vicuna-7b) & 11.78/31.78 & 0.68/1.59 & 12.54/58.89 & 4.78/87.20 & 0.61/61.15 & 14.45/47.53 & 28.48/34.69 & 37.48/55.05 & \\
BLIP-2 $^{\clubsuit}$(FlanT5-xxl) & 12.49/84.03 & 15.11/84.69 & 0.08/2.14 & 1.75/96.84 & 0.00/0.00 & 9.92/75.89 & 11.73/45.55 & 17.26/88.84 & \\
Sphinx $^{\clubsuit}$	&44.76/66.69	&22.13/79.69	&37.84/73.07	&39.88/70.62	&20.68/86.57	&83.93/53.51	&66.26/71.15	&60.66/61.43\\
Internlm $^{\clubsuit}$ &53.08/35.01&	40.78/60.05&	55.70/50.85	&57.82/45.02	&23.45/80.27	&71.21/42.34	&73.98/36.00	&83.00/19.69\\
InstructBLIP$^{\clubsuit}$ (FlanT5-xl) & 17.24/87.62 & 26.02/88.06 & 0.00/0.00 & 5.70/93.93 & 0.00/0.00 & 12.72/83.13 & 37.07/82.85 & 49.18/81.28 & \\
InstructBLIP$^{\clubsuit}$ (FlanT5-xxl) & 16.34/81.50 & 16.04/85.54 & 0.00/0.00 & 3.58/98.31 & 0.00/0.00 & 13.26/76.86 & 32.05/65.84 & 37.70/67.57 & \\
Llava-v1.5-7b$^{\clubsuit}$   & 46.81/58.13 & 23.23/77.63 & 36.56/72.97 & 38.76/66.47 & 9.80/91.56 & 63.10/54.70 & 80.14/48.06 & 69.85/32.92 \\
\rowcolor{lightblue}
w/ Graph Understanding Data  & 54.87/38.55 & 49.86/42.36 & 30.37/64.41 & 49.86/40.49 & 8.50/90.45 & 35.44/53.50 & 71.90/14.77 & 58.73/24.07 \\
Llava-v1.5-13b$^{\clubsuit}$   & 51.18/53.41 & 22.60/76.91 & 38.80/70.26 & 41.93/63.50 & 9.89/91.72 & 67.88/54.21 & 76.26/45.21 & 67.40/33.59 \\
\rowcolor{lightblue}
w/ Graph Understanding Data  & 55.76/36.09 & 47.57/38.91 & 31.47/61.66 & 50.81/35.17 & 9.77/86.36 & 54.45/56.46 & 72.07/11.80 & 60.54/14.60 \\
Qwen-Plus	&30.46/64.78	&27.42/82.37	&10.59/68.46&	6.16/81.60	&1.32/64.62	&75.93/58.65	&48.63/50.41	&33.71/60.56 \\
Qwen-max	& 25.71/63.21&	20.92/83.50&	16.70/76.00	&1.63/95.70	&1.12/96.58	&42.59/55.55&	40.47/51.61&	35.17/55.81\\
Gemini (0-shot)  & 23.26/52.35 & 21.65/80.09 & 19.11/66.94 & 16.18/83.09 & 4.79/94.78 & 66.01/53.90 & 39.40/37.80 & 40.83/52.60 \\
GPT-4V (0-shot)   & 14.10/23.09 & 17.50/72.97 & 9.64/30.58 & 23.01/66.85 & 5.31/43.62 & 24.13/32.33 & 29.22/38.03 & 46.14/42.74 \\
GPT-4V (4-shot)  & 20.63/34.52 & 26.25/69.95 & 13.19/51.75 & 23.40/61.90 & 6.12/84.94 & 46.33/51.69 & 58.49/49.79 & 48.06/35.01 \\
\hline
 & \multicolumn{9}{c}{\textit{Accuracy~$\uparrow$ on Specific Graph Theory Problems}} \\
\hline
%MiniGPT-4 $^{\clubsuit}$(Vicuna-7b, T=1.0) & 54.45 & 51.83 & 0.00 & 0.00 & 0.00 & 1.19 & 0.00 & 0.00 \\
MiniGPT-4 $^{\clubsuit}$(Vicuna-7b) & 50.67 & 48.69 & 0.00 & 0.00 & 0.00 & 5.95 & 0.00 & 0.00 \\
BLIP-2 $^{\clubsuit}$(FlanT5-xxl) & 46.63 & 61.26 & 0.00 & 0.00 & 13.79 & 0.00 & 0.00 & 0.00 \\
%mPLUG-Owl$^{\clubsuit}$ (Vicuna-7b) \\
%mPLUG-Owl$^{\clubsuit}$ (V-4-shot) \\
InstructBLIP$^{\clubsuit}$ (FlanT5-xl) & 48.79 & 47.12 & 0.00 & 0.00 & 6.90 & 0.00 & 0.00 & 0.00 \\
InstructBLIP$^{\clubsuit}$ (FlanT5-xxl) & 48.25 & 52.88 & 0.00 & 0.00 & 12.07 & 0.00 & 0.00 & 0.00 \\
Llava-v1.5-7b$^{\clubsuit}$    & 53.37 & 47.12 & 0.00 & 3.12 & 1.72 & 0.00 & 0.00 & 0.00\\
\rowcolor{lightblue}
w/ Graph Understanding Data  & 63.61~\textcolor{blue}{$\uparrow$} & 56.02~\textcolor{blue}{$\uparrow$} & 0.00 & 0.00 & 1.72 & 0.00 & 0.00 & 0.00\\
Llava-v1.5-13b$^{\clubsuit}$  & 52.83 & 47.12 & 0.00 & 4.69 & 3.45 & 0.00 & 0.00 & 0.00\\
\rowcolor{lightblue}
w/ Graph Understanding Data  & 60.38~\textcolor{blue}{$\uparrow$} & 53.93~\textcolor{blue}{$\uparrow$} & 0.00 & 0.00 & 0.00 & 4.76~\textcolor{blue}{$\uparrow$} & 3.45~\textcolor{blue}{$\uparrow$} & 0.00\\
Gemini (0-shot)  & 55.52(14.01) & 48.69(\textbf{6.80}) & 0.00 & 0.00 & 3.45 & 1.72 & 0.00 & 0.00 &\\
GPT-4V (0-shot)  & 38.81(13.74) & 49.21(0.52) & - & 3.12 & - & - & 0.00 & -\\
GPT-4V (2-shot)  & 54.98(19.13) & 52.35(0.52) & - & 6.25 & - & - & 0.00 & -\\
GPT-4V (0-COT)  & 30.45(13.20) & 50.26(0.00) & - & \textbf{7.69} & - & - & 0.00 & -\\
GPT-4V (2-COT)  & 54.71(\textbf{19.40}) & 52.87(0.52) &  -& 6.25 & - & - & 0.00 & -\\

%\\没有GPT-4V (4-shot)
\bottomrule
\end{tabular}

\end{table*}

\subsection{Evaluation Metric}

The graph theory problem evaluation will be based on three distinct sub-questions, each with its specific criteria. The first (\textit{node recognition}) assesses accuracy in counting graph nodes by comparing the answer to a standard solution. The second sub-question (\textit{edge recognition}) involves representing graph edges for four types of graphs using tuples: two-element tuples for undirected/ directed unweighted graphs; and three-element tuples including edge weights for undirected/directed weighted graphs.
The evaluation of the answer to this question encompasses two key metrics:
\begin{itemize}[leftmargin=*]
\item Correct Rate: \textit{Quantifies the proportion of correctly identified tuples in the response against the standard solution}.
\item Error Rate: \textit{Quantifies the proportion of incorrectly identified tuples relative to the total tuples in the response}.
%\item Refused Number: Reflects instances where tuples present in the standard solution are absent in the response.
%\item Half Correct Number: Reflects instances where at least half of the tuples in the solution are correctly identified in the response.
%\item Correct Number: Reflects instances where all tuples are correctly identified in the response without any errors.
\end{itemize}

%The third question (multimodal graph theory problems) varies based on the type of graph problems, including both binary (yes/no) and descriptive responses. For Connectivity and Cycle problems, evaluation includes a rough Accuracy metric for correct yes/no answers and a more rigorous Accuracy metric that requires any 'yes' responses to be accompanied by a verifiable path in the graph. used for GPT-4V and Gemini. For Topological Sort, each step of the provided solution must be correct. Shortest Path evaluations will check both the total weight of the provided shortest path and its existence within the graph. Maximum Flow problems require the precise calculation of the maximum flow value. Bipartite Graph Matching answers must correctly state the maximum number of matches and provide an error-free assignment. For Hamilton Path problems, every step of the provided path must be accurate and follow the graph's connections. Lastly, GNN problems require correct post-convolution node embeddings as per the graph convolution operation specified.

In our study, the third question about multimodal graph theory problems encompasses a range of graph problems that necessitate binary (yes/no) or descriptive responses. Specifically, for Connectivity and Cycle problems, we assess responses using two tiers of accuracy metrics: a basic metric for correct binary answers, and a more stringent metric requiring verifiable paths for affirmative responses (yes), as applied to powerful GPT-4V and Gemini models. In Topological Sort tasks, the accuracy of each step in the solution sequence is critical, where the solution is verified by the procedure. For Shortest Path problems, our evaluation criteria include verifying both the existence and the total weight of the proposed shortest path via the Python tool. In Maximum Flow problems, the focus is on accurately calculating the maximal flow value. For Bipartite Graph Matching, correct responses must identify the maximum number of matches and provide an error-free assignment. Hamilton Path challenges require each step of the proposed path to be precise and consistent with the graph's structure. Lastly, GNN tasks demand correct post-convolution node embeddings, in line with the specified graph convolution operation. Overall, we set the corresponding evaluation approach for different problems.

%This comprehensive metric ensures a thorough and precise evaluation of responses across the different types of graph theory problems, clearly distinguishing between varying degrees of correctness and providing specific metrics for assessing a wide array of answer qualities.

\subsection{Implementation Details}

Our model's training was conducted in two distinct phases. The initial phase covered 5 epochs, using a batch size of 16 and leveraging the AdamW~\cite{kingma2014adam} optimizer with a learning rate of 1e-4. Encountering a performance bottleneck after 2 epochs, we adapted the dataset by introducing a VQA task focused on incorporating fine-grained edge information. This adjustment marked the inception of the second training phase, building upon the initial training phase. The subsequent training employed modified hyperparameters, including a revised data path and an adjusted learning rate of 3e-5, with the total epochs set to 3. Throughout both phases, batch sizes, optimizer settings, and the utilization of gradient checkpoint and lazy preprocessing remained consistent. During inference, the temperature parameter is assigned a default value of 0.2, and the beam size is configured to 1. 

\begin{figure}[t]
    \centering
    \includegraphics[width=0.48\textwidth]{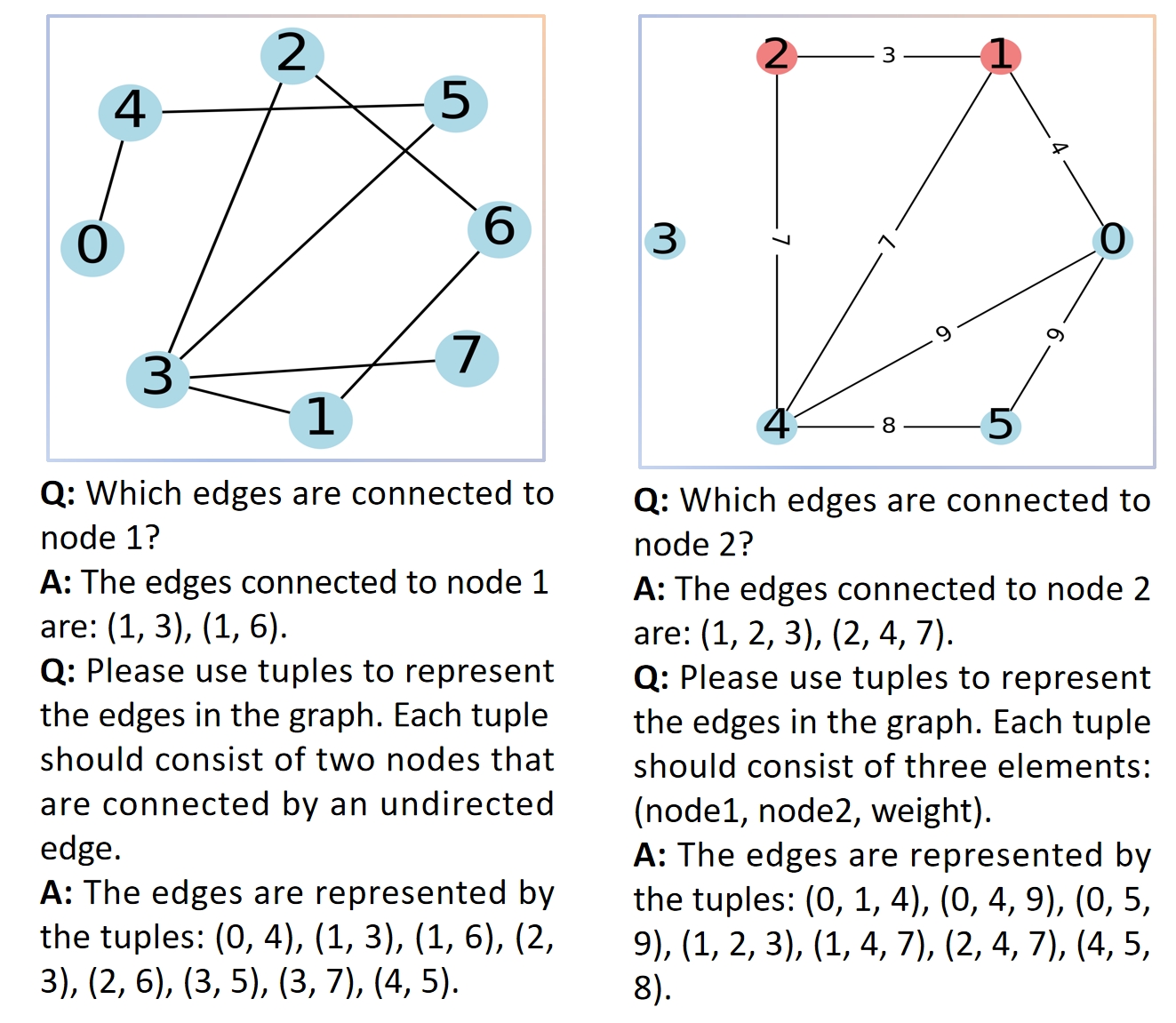}
    \caption{It illustrates two scenarios in augmented graph understanding data: 1) Overall Edge Recognition, focusing on identifying and interpreting the connections between nodes; 2) Edge-relevant VQA, which addresses questions specifically related to the visual aspects and significance of the graph's edges and nodes.}
    \label{fig:data_aug}
\end{figure}

\section{Comparative Analysis of LMMs}

In this section, we compare various supervised finetuning LMMs, Gemini, and GPT-4V on graphical structures understanding and multimodal graph theory problem-solving, especially for the effect of supervised fine-tuning approaches.

\subsection{Graphical Structures Understanding Ability}

Compared to powerful GPT-4V and Gemini, open-source LMMs such as LLava and InstructBLIP, have also achieved impressive visual understanding capability on open-world images. By analyzing the performance of GPT-4V, Gemini, and other LMMs, we can observe the overall performance and exposed problems of current LMMs on graph theory problems.
Table~\ref{tab:prompt_gpt} shows the prompting methods for LMMs.

Table~\ref{tab:initial_results} displays the results for node and edge recognition in VisionGraph. Open-source LMMs initially exhibited inferior zero-shot performances, prompting us to fine-tune them using the VisionGraph training set. This fine-tuning led to improved results, particularly notable in the Llava-v1.5-7b/13b models, which outperformed GPT-4V in both node and edge recognition tasks. However, GPT-4V still excelled over Gemini in node recognition and demonstrated a lower error rate in edge recognition, suggesting superior spatial understanding than Gemini. Additionally, GPT-4V showed improved edge recognition performance when moving from a zero-shot to a four-shot setting. Regarding edge recognition, all LMMs exhibit a notably high error rate, and the error rate outperforms the right rate. These findings suggest that while LMMs benefit from continued training on multimodal graph problems, enhancing their spatial perception, there remains significant potential for further improvement. This may be primarily due to an inferior spatial perception ability, which will significantly impact the accuracy of next-step visual reasoning.

\subsection{Effects of Supervised Fine-tuning Approaches}

After evaluating the visual recognition ability, we further analyze the graph reasoning accuracy of LMMs and the effects of supervised fine-tuning approaches. The experimental results are shown in Table~\ref{tab:initial_results}. First, we introduce 200k edge recognition and edge-relevant VQA data as shown in Figure~\ref{fig:data_aug} to enhance the graph understanding capability. They are constructed by generating the graph according to random setting nodes, edges, and questions. We observe that introducing more graph understanding will enhance the edge recognition accuracy, especially in lowering the error rate. While comparing accuracy on specific graph theory problems, we find that the data augmentation method significantly improves the performance of the model in terms of Cycle (Llava-v1.5-13b: 60.38 vs. 52.83) and Connectivity (Llava-v1.5-13b: 53.93 vs. 47.12). Hence, to improve the multimodal graph reasoning capability of LMMs, we need to introduce more low-level perception data. In addition, from Table~\ref{tab:initial_results}, we observe that the few-shot setting also improves the graph perception and reasoning accuracy of GPT-4V, which also indicates that introducing more data to improve LMMs is effective. This may be attributed to the long-tail distribution of training data in LMMs.

\section{Improving Multi-step Reasoning Capability of LMMs}

%The above results and analysis indicate that contemporary LMMs, including GPT-4V, exhibit limited proficiency in graph structure comprehension, which hampers their effectiveness in solving graph theory problems. However, GPT-4V remains the top-performing large multimodal model. 
This section mainly introduces and evaluates the Multimodal Graph Agent approach: Description Program Reasoning (DPR), designed for multimodal graph theory challenges.

\begin{figure}[t]
    \centering
    \includegraphics[width=0.48\textwidth]{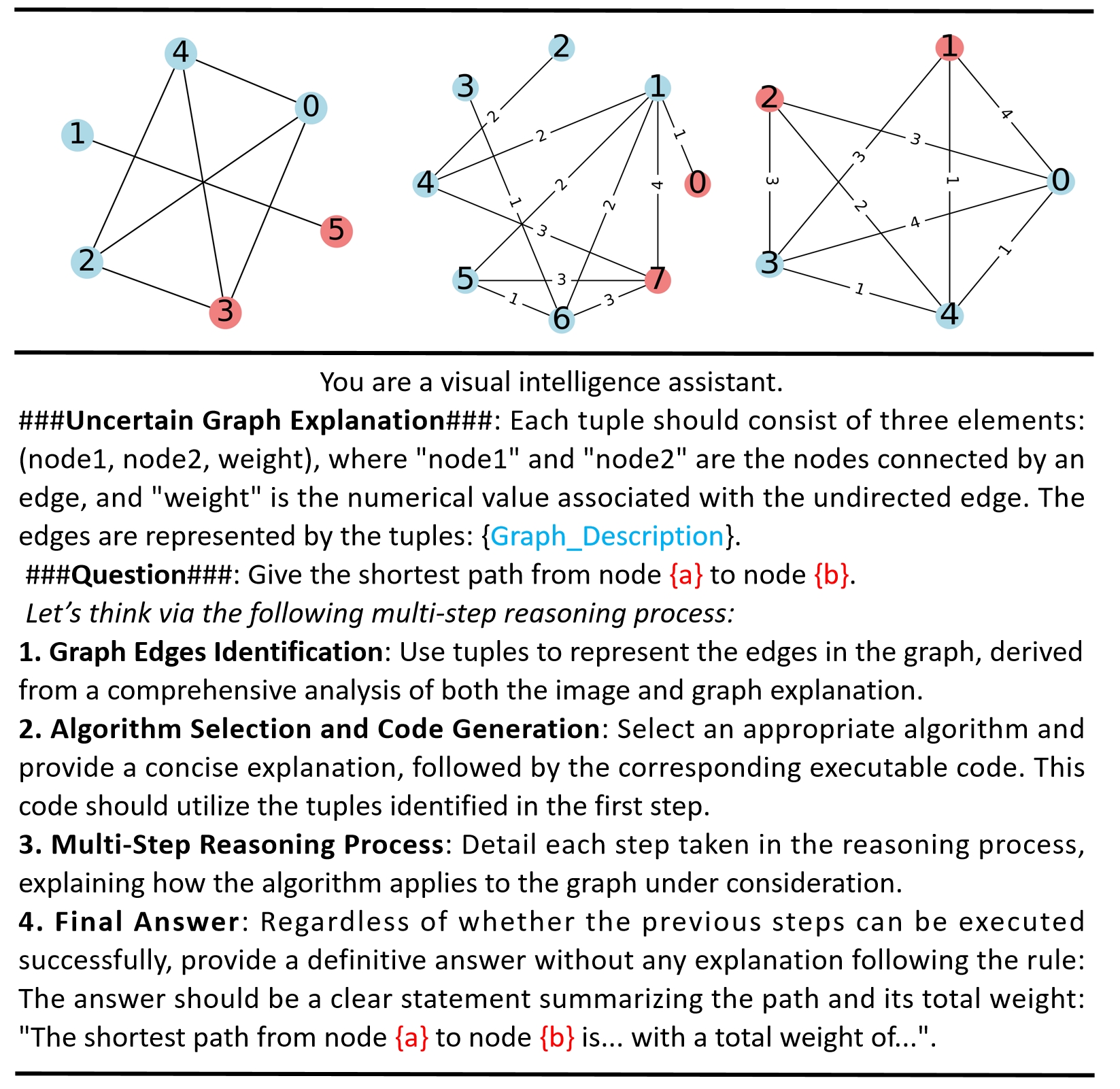}
    \caption{Overview of the prompting approach for GPT-4V (DPR).}
    \label{fig:prompt-DPR}
\end{figure}

\begin{table*}[t]
\renewcommand\arraystretch{1.15}
\tabcolsep=0.05cm
\caption{Model performance on three common graph theory problems in VisionGraph. ${\clubsuit}$ refers to that the corresponding model is trained using the training set of VisionGraph. ``w/ Python'' shows using external Python interpretation to run algorithms and return final answers. }
\label{tab:our_method_results}
\centering
\scriptsize
\begin{tabular}{l|cccc|cccc|ccc}
\toprule
Task Types $\rightarrow$ & \multicolumn{4}{c|}{Connectivity $\uparrow$} & \multicolumn{4}{c|}{Cycle $\uparrow$} & \multicolumn{3}{c}{Shortest Path $\uparrow$} \\ 
Model$\downarrow$ & Easy & Medium & Hard & Avg. & Easy & Medium & Hard  & Avg. & Easy & Hard  & Avg.\\
\hline
MiniGPT-4 $^{\clubsuit}$(Vicuna-7b) & 60.71 & 53.57 & 52.94 & 54.45 & 36.00 & 51.40 & 59.32 & 51.83 & 0.00 & 0.00 & 0.00 \\
%MiniGPT-4 $^{\clubsuit}$(Vicuna-7b, T=0.2) & 51.79 & 55.61 & 42.02 & 50.67 & 36.00 & 49.53 & 52.54 & 48.69 & 0.00 & 0.00 & 0.00 \\
BLIP-2 $^{\clubsuit}$(FlanT5-xxl) &  37.50&  43.37&  56.30&  46.63&  88.00&  63.55&  45.76&  61.26&  0.00&  0.00&  0.00\\
%mPLUG-Owl$^{\clubsuit}$ (Vicuna-7b) \\
%mPLUG-Owl$^{\clubsuit}$ (V-4-shot) \\
%InstructBLIP$^{\clubsuit}$ (Vicuna-13b) &  &  &  &  &  &  &  &  & \\
InstructBLIP$^{\clubsuit}$ (FlanT5-xl) &  46.43&  46.43&  53.78&  48.79&  36.00&  50.47&  45.76&  47.12&  0.00&  0.00&  0.00\\
Sphinx	&39.29	&45.41	&52.1&	46.63&	64.00	&49.53	&54.24	&52.88	&6.90	&0.00&	3.12\\
Sphinx$^{\clubsuit}$ w/ DPR	&67.86&	59.69&	52.94&	58.76&	64.00	&49.53	&54.24&	52.88&	13.78&	0.00&	6.25\\
Internlm$^{\clubsuit}$	&78.57&	66.33	&52.10&	52.94	& 52.00	&55.14&	59.32	&56.02&	0.00&	0.00&	0.00\\
Internlm$^{\clubsuit}$ w/ DPR & 	89.29	&72.96	&56.30	&70.08	&44.00	&57.01&	64.41&	57.59	&0.00&	0.00	&0.00\\
Llava-v1.5-7b$^{\clubsuit}$  & 64.29 & 50.00 & 53.78 & 53.27 & 36.00 & 50.47 & 45.76 & 47.12 & 6.90 & 0.00 & 3.12 \\
w/ Graph Understanding Data  & 89.29 & 64.80 & 49.58 & 63.61 & 68.00 & 53.27 & 55.93 & 56.02 & 0.00 & 0.00 & 0.00 \\
\rowcolor{lightblue}
w/ DPR & 80.36 & 68.37 & 48.74 & 63.88~\textcolor{blue}{$\uparrow$} & 68.00 & 51.40 & 55.93 & 54.97 & 0.00 & 0.00 & 0.00 \\ 
Llava-v1.5-13b$^{\clubsuit}$  & 71.43 & 49.49 & 49.58 & 52.83 & 36.00 & 50.47 & 45.76 & 47.12 & 10.34 & 0.00 & 4.69 \\
w/ Graph Understanding Data  & 78.57 & 62.76 & 47.90 & 60.38 & 64.00 & 51.40 & 54.24 & 53.93 & 0.00 & 0.00 & 0.00 \\
\rowcolor{lightblue}
w/ DPR & 83.93 & 70.92 & 50.42 & 66.31~\textcolor{blue}{$\uparrow$} & 60.00 & 64.49 & 55.93 & 61.26~\textcolor{blue}{$\uparrow$} & 0.00 & 0.00 & 0.00 \\
Gemini (0-shot)  & 69.64(39.29) & 56.63(13.78) & 47.06(2.52) & 55.52(14.01) & 60.00(0.00) & 47.66(4.67) & 45.76(13.56) & 48.69(6.80) & 0.00 & 0.00 & 0.00  \\
\rowcolor{lightblue}
Gemini (DPR)  & 66.07(57.14) & 52.04(27.04) & 36.97(5.88) & 49.32(24.79) & 76.00(16.00) & 27.10(5.61) & 22.03(0.00) & 31.93(5.23) & 0.00 & 0.00 & 0.00  \\
Qwen-plus	&62.50(19.64)&	56.63(9.69)	&47.06(3.36)&	54.45(9.16)	&64.00(0.00)	&49.53(0.00)	&54.24(0.00)&	52.88(0.00)&	0.00&	0.00	&0.00 \\
\rowcolor{lightblue}
Qwen-plus w/ DPR	&57.14(1.79)	&46.43(4.08)&	35.29(5.88)	&44.47(4.31)	&64.00(16.00)&	56.07(22.43)	&52.54(20.34)	&56.02(20.94)	&6.90	&0.00	&3.12\\
Qwen-max	&62.50(16.07)&	56.63(3.06)&	46.22(0.84)&	54.18(4.31)	&64.00(16.00)&	49.53(0.00)	&54.24(0.00)&	52.88(0.00)	&0.00	&0.00	&0.00\\
\rowcolor{lightblue}
Qwen-max w/ DPR	& 60.71(12.50)&	51.02(12.24)	&27.73(6.72)	&45.01(10.51)	&64.00(16.00)&	52.34(10.28)	&50.85(1.69)	&53.40(8.38)&	20.69	&2.86	&10.93\\

GPT-4V (0-shot)  & 69.64(46.43) & 42.86(12.76) & 17.65(0.00) & 38.81(13.74) & 60.00(4.00) & 48.60(0.00) & 45.76(0.00) & 49.21(0.52) & 6.90 & 0.00 & 3.12 \\
GPT-4V (2-shot)  & 67.86(42.86) & 56.12(18.88) & 47.06(8.40) & 54.98(19.13) & 64.00(4.00) & 48.60(0.00) & 54.24(0.00) & 52.35(0.52) & 13.79 & 0.00 & 6.25 \\
GPT-4V (0-COT)  & 64.29(37.50) & 34.69(13.78) & 7.56(0.84) & 30.45(13.20) & 64.00(0.00) & 47.66(0.00) & 49.15(0.00) & 50.26(0.00) & 17.24 & 0.00 & 7.69 \\
GPT-4V (2-COT)  & 67.86(44.64) & 56.63(22.96) & 45.38(1.68) & 54.71(19.40) & 64.00(4.00) & 49.53(0.00) & 54.24(0.00) & 52.87(0.52) & 13.79 & 0.00 & 6.25  \\
GPT-4V (DPR) & 92.86(89.29) & 58.67(44.90) & 36.97(16.81) & 56.87(42.58) & 76.00(52.00) & 48.60(15.89) & 45.76(1.69) & 51.30(16.23) & 24.14 & 2.86 & 12.50 \\
\rowcolor{lightblue}
w/ Python & 92.86(91.07) & 61.73(53.06) & 51.26(35.29) & \textbf{63.07(53.09)}~\textcolor{blue}{$\uparrow$} & 88.00(72.00) & 61.68(34.58) & 55.93(20.34) & \textbf{63.35(35.07)}~\textcolor{blue}{$\uparrow$} & 31.03 & 11.43 & \textbf{20.31}~\textcolor{blue}{$\uparrow$} \\

%\\没有GPT-4V (4-shot)
\bottomrule
\end{tabular}

\end{table*}

\subsection{Description Program Reasoning}

To improve the performance of GPT-4V on multimodal graph theory problems, inspired by LLMs-powered Agents~\cite{wang2023survey_agents,qin2023toolllm,yao2022react}, we devise a multimodal graph problem-solving approach: Description-Program-Reasoning (DPR), which is a natural language and code interleaved reasoning chain. Specifically, we first use the Llava-7b augmented by the graph understanding data to produce high-quality graph explanations, which are fed into GPT-4V to enhance the graphical structure understanding. As the process shown in Figure~\ref{fig:prompt-DPR}, we prompt GPT-4V to answer the specific graph problems based on the visual graph and its descriptions:

\begin{itemize}[leftmargin=*]
    \item Create the Adjacency Matrix of the graph according to the visual graph and its description generated by small models. We require the adjacency matrix to be given in the form of triples, i.e., (node1, node2, weight) for directed graphs and (node1, node2) for undirected graphs. 
    \item Select the appropriate graph theory algorithm and generate the corresponding codes. This step aims to generate codes used for performing multi-step reasoning on the visual graph.
    \item Perform multi-step reasoning based on the Adjacency Matrix and generated codes. We can also use an external Python interpreter to perform multi-step reasoning, enhancing the overall performance.
    \item Give the final answer in the demanding format.
\end{itemize}

\begin{figure}[t]
    \centering
    \includegraphics[width=0.48\textwidth]{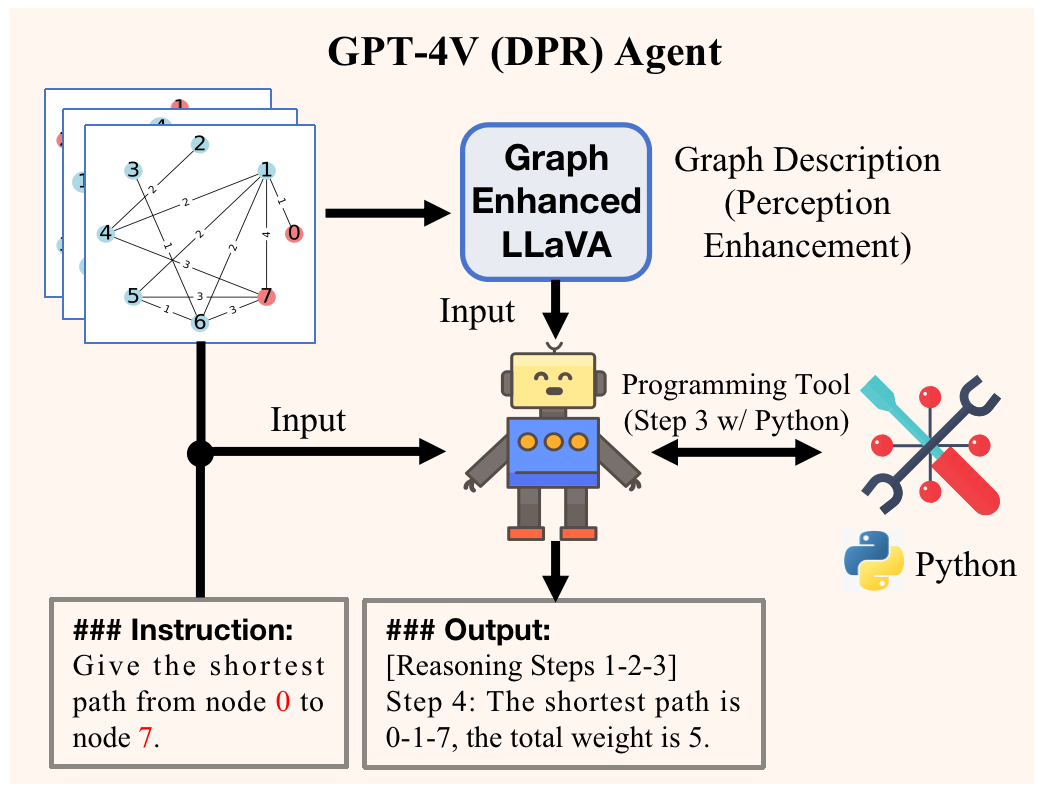}
    \caption{An Illustration of the designed GPT-4V (DPR) Agent.}
    \label{fig:model-DPR}
\end{figure}

Hence, \textit{GPT-4V with DPR is a comprehensive multi-modal Agent that integrates complex task decomposition, small model enhancement, code generation, and tool invocation}. We illustrate the proposed DPR in Figure~\ref{fig:prompt-DPR} and \ref{fig:model-DPR}. In the following, we employ the DPR prompting and evaluate it on three representative tasks in the VisionGraph benchmark with varying difficulties. 

\subsection{Results and Analysis}

\textbf{Ablation Study}.
Table~\ref{tab:our_method_results} shows all LMMs' performances on Connectivity, Cycle, and Shortest Path with varying complexity. 
Since Llava can not follow instructions (shown in Figure~\ref{fig:prompt-DPR}) to generate the detailed reasoning process, we apply DPR to Llava by adding the results of edge recognition to the conversation history so that its graph description can be attention when answering the graph theory question. The whole process could be simplified as ``Description-Reasoning''. 
We observe that DPR significantly improves the multi-step reasoning capabilities of LMMs and performs better on more complex Shortest Paths. For GPT-4V, DPR equipped with the Python tool exceeds its 2-shot performance: gain by 8\%, 10\%, and 14\% on Connectivity, Cycle, and Shortest Path, respectively. It indicates DPR shows its superiority in complex multi-step graph reasoning.
However, we also observe that GPT-4V and Gemini gain a low performance on hard graph theory problems (nodes $>$ 20 or nodes $>$ 11 on Shortest Path). Especially on the Shortest Path, many LMMs are not capable of solving this task, such as 0.0 for Llava-v1.5-7b/13b and Gemini. Overall, we demonstrate that: \textit{1) The natural language and code interactive reasoning chain enhances complex multi-modal reasoning capabilities of LMMs. 2) The limited graph perception inherent in LMMs results in their poor performance in multi-step reasoning on the visual graph; 3) GPT-4V exhibits superior multi-step graph reasoning abilities compared to Gemini.}

%As shown in Table~\ref{tab:our_method_results}, DPR enhances the overall performance of GPT-4V and Llava-13b;
%which is substantially below human performance

\begin{figure}
    \centering
    \includegraphics[width=0.48\textwidth]{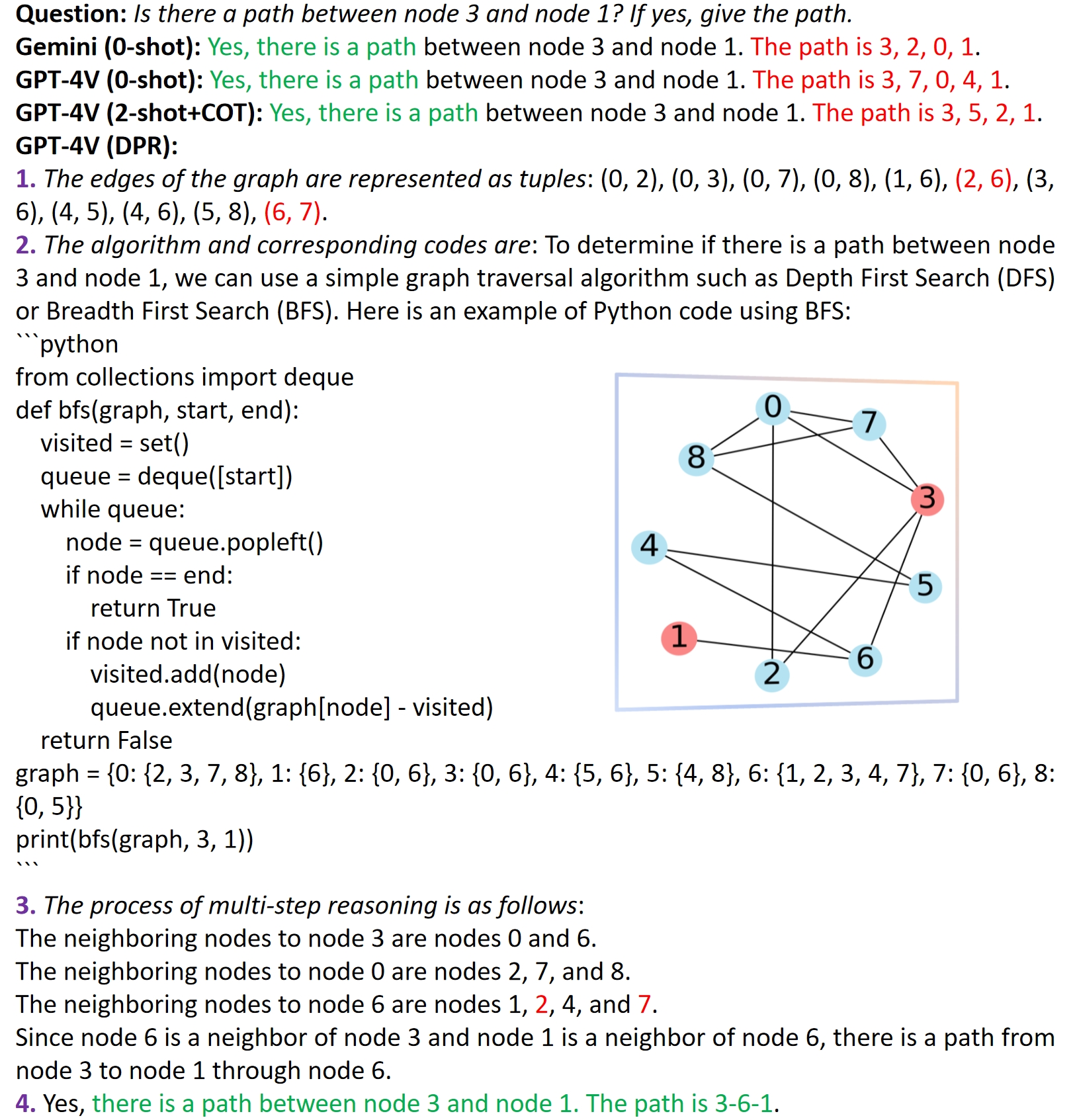}
    \caption{A case (Connectivity) illustrates results generated by different models. \textcolor{red}{Red} and \textcolor{green}{green} words are incorrect and correct contents, respectively.}
    \label{fig:cases_study}
\end{figure}

\noindent \textbf{Case Study}.
We report a representative case in Figure~\ref{fig:cases_study} to analyze the performance of LMMs. Gemini Pro and GPT-4V made the correct assessment ``There is a path between node 3 and node 1.'' More various cases (Five cases shown in Figures~\ref{fig:cases_study_shortest}-\ref{fig:cases_5}) are shown in \textit{Appendix}.  However, when asked to specify the correct path, only GPT-4V (DPR) provided an accurate answer. The other samples erroneously constructed paths using edges that do not exist in the graph. This may be attributed to the fact that most LMMs have relatively weaker graph recognition capabilities. We suggest supplementing these models with highly accurate graphical interpretation information (in this case, the edge information provided had an accuracy rate of 81.81\%) could compensate for GPT-4V's shortcomings of graph recognition. Additionally, GPT-4V shows proficiency in accurately completing tasks involving the selection of algorithms for specific graph theory problems and utilizing the algorithm for reasoning analysis. To conclude, the proposed DPR can assist GPT-4V in 
maximizing its inherent strengths and effectively mitigating its weaknesses in graph theory problems.

\section{Conclusion}

In this study, we delved into the capabilities of large multimodal models (LMMs) in addressing multimodal graph theory challenges. Initially, we developed the VisionGraph benchmark, tailored for evaluating LMMs. This benchmark not only encompasses node and edge identification tasks to gauge the graphical comprehension of LMMs but also incorporates eight distinct graph theory problems to test their multi-step reasoning abilities. Subsequently, we conducted a comprehensive analysis of various LMMs, including GPT-4V and Gemini, using VisionGraph. This analysis focused on two key aspects: the understanding of graphical structures and the impact of supervised fine-tuning methods. Furthermore, we introduced a multimodal graph theory-oriented Agent, named Description Programming Reasoning (DPR). DPR is uniquely designed to integrate intricate task decomposition, perception enhancement attuned to smaller models, advanced code generation, and the utilization of external tools. Through experimental evaluations, it has been demonstrated that DPR significantly enhances the performance of LMMs on multimodal graph theory tasks. 
%The experimental insights have paved the way for our future research direction, which will concentrate on further improving the graphical structure perception and multi-step reasoning capabilities of LMMs, specifically through advanced interleaved reasoning strategies that integrate natural language and programming languages. 

%In this work, we explore the large multimodal models to solve multimodal graph theory problems. First, we constructed a multimodal graph theory evaluation benchmark VisionGraph, which includes node and edge identification problems used to assess the graphical understanding ability of LMMs. It also contains eight types of graph theory problems to evaluate their multi-step reasoning capabilities. Then, we analyzed GPT-4V, Gemini, and other open-source LMMs on VisionGraph in terms of Graphical Structures Understanding Ability and Effects of Supervised Fine-tuning approach. Finally, we designed a multimodal graph theory-oriented Agent: Description Programming Reasoning (DPR), which integrates complex task decomposition, small model-aware perception enhancement, code generation, and the use of external tools. Experimental results show that DPR significantly improves the performance of LMMs on multimodal graph theory problems. The experimental insights have paved the way for our future research direction, which will concentrate on further improving the graphical structure perception and multi-step reasoning capabilities of LMMs, specifically through advanced interleaved reasoning strategies based on natural language and program languages. 

\section*{Acknowledge}
Thanks for the efforts from reviewers and action editors. This work is supported by grants: Natural Science Foundation of China (No. 62376067).

\section*{Impact Statements}
Our paper mainly presents a multimodal graph theory problem benchmark named VisionGraph to check the multimodal planning performance of LMMs. We also introduce a corresponding multimodal agent to handle such complex visual maths. This research marks a significant contribution to several fields: it advances the spatial perception capabilities of Large Multimodal Models, propels the frontiers of Visual Intelligence in areas such as visual math and robotics planning, supports the application of AI in scientific endeavors, particularly in Mathematics, and offers insights into Strategy Optimization for industrial applications.

Ethical Considerations:
\begin{itemize}
    \item \textbf{Data Source and Tools}. The foundational data for our benchmark originates from publicly accessible text datasets, explicitly designated for academic research purposes. This ensures compliance with data usage norms and ethical standards in academic contexts. Additionally, the methodology employed for data construction exclusively utilizes open-source graph construction tools. Emphasizing transparency and community engagement, we commit to making our benchmark openly available, facilitating further research and validation by the academic community.
    \item \textbf{Model Usage and Bias Acknowledgement}. Our research incorporates the use of two commercial LMMs: GPT-4V and Gemini Pro. It's crucial to acknowledge that the content generated by these models may inherently carry biases. This is a reflection of the models' training datasets and algorithms, rather than a deliberate design choice. Recognizing this, we approach our findings with a critical perspective and urge users to consider potential biases when interpreting the results.
\end{itemize}

\bibliography{example_paper}

\begin{thebibliography}{28}
\providecommand{\natexlab}[1]{#1}
\providecommand{\url}[1]{\texttt{#1}}
\expandafter\ifx\csname urlstyle\endcsname\relax
  \providecommand{\doi}[1]{doi: #1}\else
  \providecommand{\doi}{doi: \begingroup \urlstyle{rm}\Url}\fi

\bibitem[Anderson et~al.(2018)Anderson, Wu, Teney, Bruce, Johnson, S{\"u}nderhauf, Reid, Gould, and Van Den~Hengel]{anderson2018vision}
Anderson, P., Wu, Q., Teney, D., Bruce, J., Johnson, M., S{\"u}nderhauf, N., Reid, I., Gould, S., and Van Den~Hengel, A.
\newblock Vision-and-language navigation: Interpreting visually-grounded navigation instructions in real environments.
\newblock In \emph{Proceedings of the IEEE conference on computer vision and pattern recognition}, pp.\  3674--3683, 2018.

\bibitem[Bai et~al.(2023)Bai, Bai, Yang, Wang, Tan, Wang, Lin, Zhou, and Zhou]{bai2023qwen}
Bai, J., Bai, S., Yang, S., Wang, S., Tan, S., Wang, P., Lin, J., Zhou, C., and Zhou, J.
\newblock Qwen-vl: A frontier large vision-language model with versatile abilities.
\newblock \emph{arXiv preprint arXiv:2308.12966}, 2023.

\bibitem[Chen et~al.(2024)Chen, Pitawela, Zhao, Zhou, Chen, and Wu]{chen2024webvln}
Chen, Q., Pitawela, D., Zhao, C., Zhou, G., Chen, H.-T., and Wu, Q.
\newblock Webvln: Vision-and-language navigation on websites.
\newblock In \emph{Proceedings of the AAAI Conference on Artificial Intelligence}, volume~38, pp.\  1165--1173, 2024.

\bibitem[Dai et~al.(2023)Dai, Xiong, and Ku]{dai2023llm}
Dai, S.-C., Xiong, A., and Ku, L.-W.
\newblock Llm-in-the-loop: Leveraging large language model for thematic analysis.
\newblock \emph{arXiv preprint arXiv:2310.15100}, 2023.

\bibitem[Ektefaie et~al.(2022)Ektefaie, Dasoulas, Noori, Farhat, and Zitnik]{ektefaie2022geometric}
Ektefaie, Y., Dasoulas, G., Noori, A., Farhat, M., and Zitnik, M.
\newblock Geometric multimodal representation learning.
\newblock \emph{arXiv preprint arXiv:2209.03299}, 2022.

\bibitem[Gu et~al.(2022)Gu, Stefani, Wu, Thomason, and Wang]{gu2022vision}
Gu, J., Stefani, E., Wu, Q., Thomason, J., and Wang, X.~E.
\newblock Vision-and-language navigation: A survey of tasks, methods, and future directions.
\newblock \emph{arXiv preprint arXiv:2203.12667}, 2022.

\bibitem[Imani et~al.(2023)Imani, Du, and Shrivastava]{imani2023mathprompter}
Imani, S., Du, L., and Shrivastava, H.
\newblock Mathprompter: Mathematical reasoning using large language models.
\newblock \emph{arXiv preprint arXiv:2303.05398}, 2023.

\bibitem[Kingma \& Ba(2014)Kingma and Ba]{kingma2014adam}
Kingma, D.~P. and Ba, J.
\newblock Adam: A method for stochastic optimization.
\newblock \emph{arXiv preprint arXiv:1412.6980}, 2014.

\bibitem[Li et~al.(2023{\natexlab{a}})Li, Li, Savarese, and Hoi]{li2023blip2}
Li, J., Li, D., Savarese, S., and Hoi, S.
\newblock Blip-2: Bootstrapping language-image pre-training with frozen image encoders and large language models.
\newblock \emph{ICML}, 2023{\natexlab{a}}.

\bibitem[Li et~al.(2023{\natexlab{b}})Li, Hu, Chen, Ma, and Zhang]{li2023lmeye}
Li, Y., Hu, B., Chen, X., Ma, L., and Zhang, M.
\newblock Lmeye: An interactive perception network for large language models.
\newblock \emph{arXiv preprint arXiv:2305.03701}, 2023{\natexlab{b}}.

\bibitem[Li et~al.(2023{\natexlab{c}})Li, Liu, Wang, Liang, Liu, Wang, Cui, Tu, Wang, and Zhou]{li2023comprehensive_medical}
Li, Y., Liu, Y., Wang, Z., Liang, X., Liu, L., Wang, L., Cui, L., Tu, Z., Wang, L., and Zhou, L.
\newblock A comprehensive study of gpt-4v's multimodal capabilities in medical imaging.
\newblock \emph{medRxiv}, pp.\  2023--11, 2023{\natexlab{c}}.

\bibitem[Li et~al.(2023{\natexlab{d}})Li, Wang, Hu, Chen, Zhong, Lyu, and Zhang]{li2023comprehensive}
Li, Y., Wang, L., Hu, B., Chen, X., Zhong, W., Lyu, C., and Zhang, M.
\newblock A comprehensive evaluation of gpt-4v on knowledge-intensive visual question answering.
\newblock \emph{arXiv preprint arXiv:2311.07536}, 2023{\natexlab{d}}.

\bibitem[Lin et~al.(2023)Lin, Liu, Zhang, Gao, Qiu, Xiao, Qiu, Lin, Shao, Chen, et~al.]{lin2023sphinx}
Lin, Z., Liu, C., Zhang, R., Gao, P., Qiu, L., Xiao, H., Qiu, H., Lin, C., Shao, W., Chen, K., et~al.
\newblock Sphinx: The joint mixing of weights, tasks, and visual embeddings for multi-modal large language models.
\newblock \emph{arXiv preprint arXiv:2311.07575}, 2023.

\bibitem[Liu et~al.(2023)Liu, Li, Wu, and Lee]{liu2023visual}
Liu, H., Li, C., Wu, Q., and Lee, Y.~J.
\newblock Visual instruction tuning.
\newblock \emph{NeurIPS}, 2023.

\bibitem[Lu et~al.(2023)Lu, Bansal, Xia, Liu, Li, Hajishirzi, Cheng, Chang, Galley, and Gao]{lu2023mathvista}
Lu, P., Bansal, H., Xia, T., Liu, J., Li, C., Hajishirzi, H., Cheng, H., Chang, K.-W., Galley, M., and Gao, J.
\newblock Mathvista: Evaluating mathematical reasoning of foundation models in visual contexts.
\newblock \emph{arXiv preprint arXiv:2310.02255}, 2023.

\bibitem[Luo et~al.(2023)Luo, Sun, Xu, Zhao, Lou, Tao, Geng, Lin, Chen, and Zhang]{luo2023wizardmath}
Luo, H., Sun, Q., Xu, C., Zhao, P., Lou, J., Tao, C., Geng, X., Lin, Q., Chen, S., and Zhang, D.
\newblock Wizardmath: Empowering mathematical reasoning for large language models via reinforced evol-instruct.
\newblock \emph{arXiv preprint arXiv:2308.09583}, 2023.

\bibitem[OpenAI(2023)]{gpt4}
OpenAI.
\newblock Gpt-4 technical report.
\newblock \emph{https://arxiv.org/abs/2303.08774}, 2023.

\bibitem[Qin et~al.(2023)Qin, Liang, Ye, Zhu, Yan, Lu, Lin, Cong, Tang, Qian, et~al.]{qin2023toolllm}
Qin, Y., Liang, S., Ye, Y., Zhu, K., Yan, L., Lu, Y., Lin, Y., Cong, X., Tang, X., Qian, B., et~al.
\newblock Toolllm: Facilitating large language models to master 16000+ real-world apis.
\newblock \emph{arXiv preprint arXiv:2307.16789}, 2023.

\bibitem[Team et~al.(2023)Team, Anil, Borgeaud, Wu, Alayrac, Yu, Soricut, Schalkwyk, Dai, Hauth, et~al.]{team2023gemini}
Team, G., Anil, R., Borgeaud, S., Wu, Y., Alayrac, J.-B., Yu, J., Soricut, R., Schalkwyk, J., Dai, A.~M., Hauth, A., et~al.
\newblock Gemini: a family of highly capable multimodal models.
\newblock \emph{arXiv preprint arXiv:2312.11805}, 2023.

\bibitem[Wake et~al.(2023)Wake, Kanehira, Sasabuchi, Takamatsu, and Ikeuchi]{wake2023gpt_robotics}
Wake, N., Kanehira, A., Sasabuchi, K., Takamatsu, J., and Ikeuchi, K.
\newblock Gpt-4v (ision) for robotics: Multimodal task planning from human demonstration.
\newblock \emph{arXiv preprint arXiv:2311.12015}, 2023.

\bibitem[Wang et~al.(2023{\natexlab{a}})Wang, Feng, He, Tan, Han, and Tsvetkov]{wang2023can}
Wang, H., Feng, S., He, T., Tan, Z., Han, X., and Tsvetkov, Y.
\newblock Can language models solve graph problems in natural language?
\newblock \emph{NeurIPS}, 2023{\natexlab{a}}.

\bibitem[Wang et~al.(2023{\natexlab{b}})Wang, Ma, Feng, Zhang, Yang, Zhang, Chen, Tang, Chen, Lin, et~al.]{wang2023survey_agents}
Wang, L., Ma, C., Feng, X., Zhang, Z., Yang, H., Zhang, J., Chen, Z., Tang, J., Chen, X., Lin, Y., et~al.
\newblock A survey on large language model based autonomous agents.
\newblock \emph{arXiv preprint arXiv:2308.11432}, 2023{\natexlab{b}}.

\bibitem[Wang et~al.(2022)Wang, Kordi, Mishra, Liu, Smith, Khashabi, and Hajishirzi]{selfinstruct}
Wang, Y., Kordi, Y., Mishra, S., Liu, A., Smith, N.~A., Khashabi, D., and Hajishirzi, H.
\newblock Self-instruct: Aligning language model with self generated instructions, 2022.

\bibitem[Wen et~al.(2023)Wen, Yang, Fu, Wang, Cai, Li, Ma, Li, Xu, Shang, et~al.]{wen2023road}
Wen, L., Yang, X., Fu, D., Wang, X., Cai, P., Li, X., Ma, T., Li, Y., Xu, L., Shang, D., et~al.
\newblock On the road with gpt-4v (ision): Early explorations of visual-language model on autonomous driving.
\newblock \emph{arXiv preprint arXiv:2311.05332}, 2023.

\bibitem[Yao et~al.(2022)Yao, Zhao, Yu, Du, Shafran, Narasimhan, and Cao]{yao2022react}
Yao, S., Zhao, J., Yu, D., Du, N., Shafran, I., Narasimhan, K., and Cao, Y.
\newblock React: Synergizing reasoning and acting in language models.
\newblock \emph{arXiv preprint arXiv:2210.03629}, 2022.

\bibitem[Zhang et~al.(2023{\natexlab{a}})Zhang, Collins, Weller, and Tenenbaum]{zhang2023ai}
Zhang, C.~E., Collins, K.~M., Weller, A., and Tenenbaum, J.~B.
\newblock Ai for mathematics: A cognitive science perspective.
\newblock \emph{arXiv preprint arXiv:2310.13021}, 2023{\natexlab{a}}.

\bibitem[Zhang et~al.(2023{\natexlab{b}})Zhang, Wang, Cao, Xu, Ouyang, Zhao, Ding, Zhang, Duan, Yan, et~al.]{zhang2023internlm}
Zhang, P., Wang, X. D.~B., Cao, Y., Xu, C., Ouyang, L., Zhao, Z., Ding, S., Zhang, S., Duan, H., Yan, H., et~al.
\newblock Internlm-xcomposer: A vision-language large model for advanced text-image comprehension and composition.
\newblock \emph{arXiv preprint arXiv:2309.15112}, 2023{\natexlab{b}}.

\bibitem[Zhu et~al.(2023)Zhu, Chen, Shen, Li, and Elhoseiny]{zhu2023minigpt}
Zhu, D., Chen, J., Shen, X., Li, X., and Elhoseiny, M.
\newblock Minigpt-4: Enhancing vision-language understanding with advanced large language models.
\newblock \emph{arXiv preprint arXiv:2304.10592}, 2023.

\end{thebibliography}
\bibliographystyle{icml2024}

%%%%%%%%%%%%%%%%%%%%%%%%%%%%%%%%%%%%%%%%%%%%%%%%%%%%%%%%%%%%%%%%%%%%%%%%%%%%%%%
%%%%%%%%%%%%%%%%%%%%%%%%%%%%%%%%%%%%%%%%%%%%%%%%%%%%%%%%%%%%%%%%%%%%%%%%%%%%%%%
% APPENDIX
%%%%%%%%%%%%%%%%%%%%%%%%%%%%%%%%%%%%%%%%%%%%%%%%%%%%%%%%%%%%%%%%%%%%%%%%%%%%%%%
%%%%%%%%%%%%%%%%%%%%%%%%%%%%%%%%%%%%%%%%%%%%%%%%%%%%%%%%%%%%%%%%%%%%%%%%%%%%%%%
\newpage
\appendix
\onecolumn
%\section{You \emph{can} have an appendix here.}

%You can have as much text here as you want. The main body must be at most $8$ pages long. For the final version, one more page can be added. If you want, you can use an appendix like this one.  
%The $\mathtt{\backslash onecolumn}$ command above can be kept in place if you prefer a one-column appendix, or can be removed if you prefer a two-column appendix.  Apart from this possible change, the style (font size, spacing, margins, page numbering, etc.) should be kept the same as the main body.
%%%%%%%%%%%%%%%%%%%%%%%%%%%%%%%%%%%%%%%%%%%%%%%%%%%%%%%%%%%%%%%%%%%%%%%%%%%%%%%
%%%%%%%%%%%%%%%%%%%%%%%%%%%%%%%%%%%%%%%%%%%%%%%%%%%%%%%%%%%%%%%%%%%%%%%%%%%%%%%

\section{Limitations}

Our work also contains some main limitations:
\begin{itemize}[leftmargin=*]
    \item \textbf{Data Distribution Imbalance of VisionGraph}.
    In our benchmark, we have included eight distinct types of multimodal graph theory problems. One of the primary challenges we encountered was the inherent difficulty in constructing diverse instances for these different types of graph theory problems. This has resulted in an uneven distribution of training and testing datasets across the various problem types. Despite this imbalance, concerted efforts were made to ensure the dataset's diversity and to cover four types of graphs, aiming for a comprehensive evaluation of LMMS' capabilities in solving graph theory problems. Moreover, our experiments indicated that augmenting the volume of training data can substantially enhance the models' spatial understanding of graphs. Future research could focus on developing more balanced datasets to further validate and extend our conclusions.
    \item \textbf{Visual Graph Construction}.
    To maintain legibility, we've set a cap on the number of nodes in undirected graphs at 25 and directed graphs at 20 during data augmentation. This restriction means our dataset doesn't encompass more complex graph structures, potentially leading to a trained model that can not recognize intricate graphs. Additionally, all nodes and edges in these graphs are randomly generated, which could result in a few duplicate graphs within the dataset, yet the questions are different. 
    \item \textbf{Model Updates and Reproducibility}.
   When the parameters of the GPT-4V and Gemini models change, current evaluations might become outdated or imprecise. To address this, our paper focuses on ensuring the reproducibility of findings: 1) We will provide all LMMs outputs under various prompt settings, allowing reliable replication and validation of our results. 2) We will release the codes and benchmarks used for research in the community.

\end{itemize}

\begin{table*}[t]
\renewcommand\arraystretch{0.90}
\footnotesize
    \centering
    \begin{tabular}{c|p{13.5cm}}
    \toprule
       Types  &  Specific Output Demands\\
       \hline
        Connect & Q2:\newline \{specific answer format requirements\}: Please use tuples to represent the edges in the graph. Each tuple should consist of two nodes that are connected by an undirected edge.\newline Q3:\newline \{the concrete problem\}: Is there a path between node a and node b in the graph?\newline \{specific answer format requirements\}: If there is a path between node a and node b, conclude your answer with 'Yes, there is a path between node a and node b. The path is... ', and provide the specific nodes involved in the path in sequence. If no path exists, please conclude with 'No, there is no path between node a and node b.' \\
        \hline
        Cycle & Q2:\newline \{specific answer format requirements\}: Please use tuples to represent the edges in the graph. Each tuple should consist of two nodes that are connected by an undirected edge.\newline Q3:\newline \{the concrete problem\}: Is there a cycle in the graph?\newline \{specific answer format requirements\}: If there is a cycle, conclude your answer with 'Yes, there is a cycle in the graph. The cycle is... ', and provide the specific nodes involved in the cycle in sequence. If no cycle exists, please conclude with 'No, there is no cycle in the graph.' \\
        \hline
        Topo. Sort & Q2:\newline \{specific answer format requirements\}: Please use tuples to represent the edges in the graph. Each tuple should consist of two nodes representing a directed edge, with the first node being the source and the second node being the destination.\newline Q3:\newline \{the concrete problem\}: Can all the nodes be visited? Give the solution. \\
        \hline
        Shortest Path & Q2:\newline \{specific answer format requirements\}: Please use tuples to represent the edges in the graph. Each tuple should consist of three elements: (node1, node2, weight), where 'node1' and 'node2' are the nodes connected by an edge, and 'weight' is the numerical value associated with the undirected edge.\newline Q3:\newline \{the concrete problem\}: Give the shortest path from node a to node b in the graph and the final answer contains the path and total weights.\newline \{specific answer format requirements\}: Please conclude your answer with a clear statement summarizing the path and its total weight, for example, 'In conclusion, the shortest path from node a to node b is... with a total weight of...'. \\
        \hline
        Max. Flow & Q2:\newline \{specific answer format requirements\}: Please use tuples to represent the edges in the graph. Each tuple should consist of three elements: <source, destination, weight>, where 'source' is the source node, 'destination' is the destination node, and 'weight' is the numerical value associated with the directed edge.\newline Q3:\newline \{the concrete problem\}: What is the maximum flow from node a to node b? \\
        \hline
        Bipartite Graph & Q2:\newline \{specific answer format requirements\}: Please use tuples to represent the edges in the graph. Each tuple should consist of two nodes that are connected by an undirected edge.\newline Q3:\newline \{the concrete problem\}: Find an assignment of jobs to applicants in such that the maximum number of applicants find the job they are interested in. \\
        \hline
        Hamilton Path & Q2:\newline \{specific answer format requirements\}: Please use tuples to represent the edges in the graph. Each tuple should consist of two nodes that are connected by an undirected edge.\newline Q3:\newline \{the concrete problem\}: Is there a path in this graph that visits every node exactly once?\newline \{specific answer format requirements\}: If there is a Hamilton path, conclude your answer with "Yes, there is a Hamilton path in the graph. The Hamilton path is ... ." If no Hamilton path exits, please conclude with "No, there is no Hamilton path in the graph." \\
        \hline
        GNNs & Q2:\newline \{specific answer format requirements\}: Please use tuples to represent the edges in the graph. Each tuple should consist of two nodes that are connected by an undirected edge.\newline Q3:\newline \{the concrete problem\}: Each node is initially assigned an embedding vector in the graph, and the embedding will be updated by the sum of its neighbors' embeddings in a simple graph convolution layer. What's the embedding of each node after one layer of simple graph convolution layer? \\
        \hline
        ALL & \{Few-shot preffix\}: The first URL refers to the first image and the second URL refers to the second image. Following the N in-context examples shown in the first composite image, answer the following question for the second image.\\
    \bottomrule
    \end{tabular}
    \caption{The explanation of specific output demands used in our prompts. Note: Q2 is about edge identification, while Q3 is about specific graph theory problems. }
    \label{tab:prompt_gpt_specific}
\end{table*}
\newpage
\section{More Cases and Detailed Prompting Technical}

\begin{figure*}[h]
    \centering
    \includegraphics[width=0.97\textwidth]{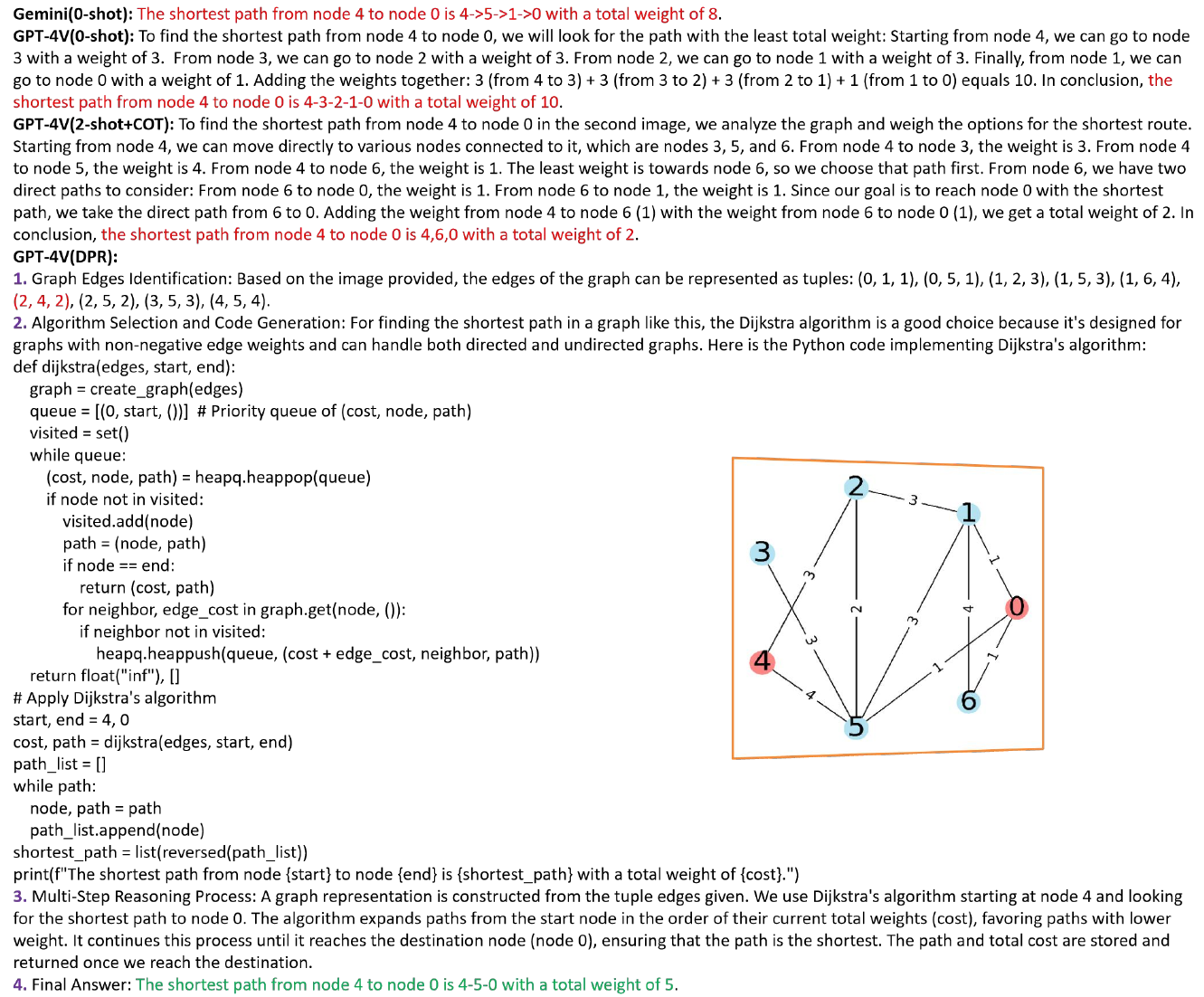}
    \caption{A case (Shortest Path) illustrates results generated by different models. \textcolor{red}{Red} and \textcolor{green}{green} words are incorrect and correct contents, respectively.}
    \label{fig:cases_study_shortest}
\end{figure*}

\begin{figure*}[t]
    \centering
    \includegraphics[width=0.85\textwidth]{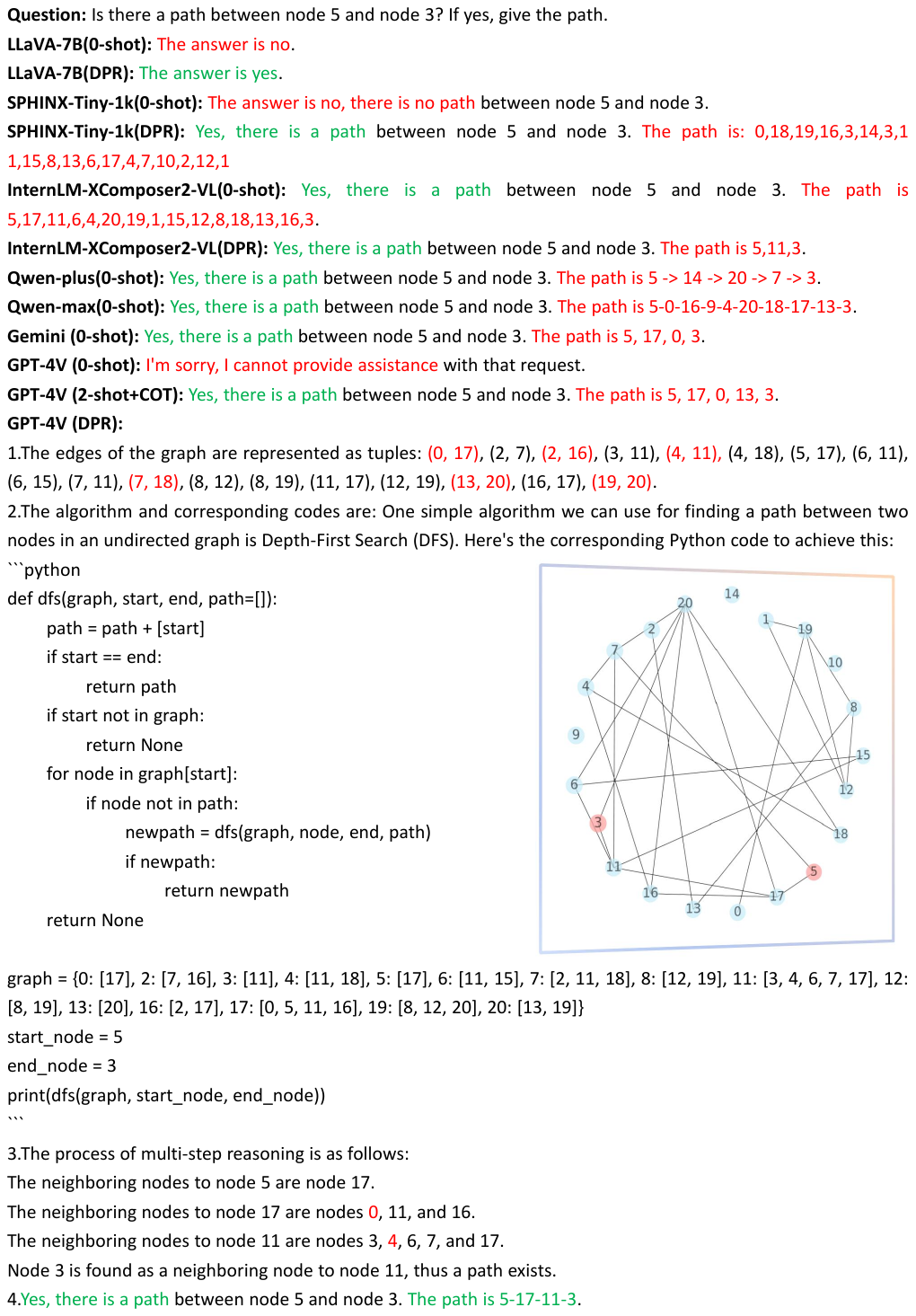}
    \caption{A case (Connectivity) illustrates results generated by different models. \textcolor{red}{Red} and \textcolor{green}{green} words are incorrect and correct contents, respectively.}
    \label{fig:cases_study_2}
\end{figure*}
\begin{figure*}[t]
    \centering
    \includegraphics[width=0.88\textwidth]{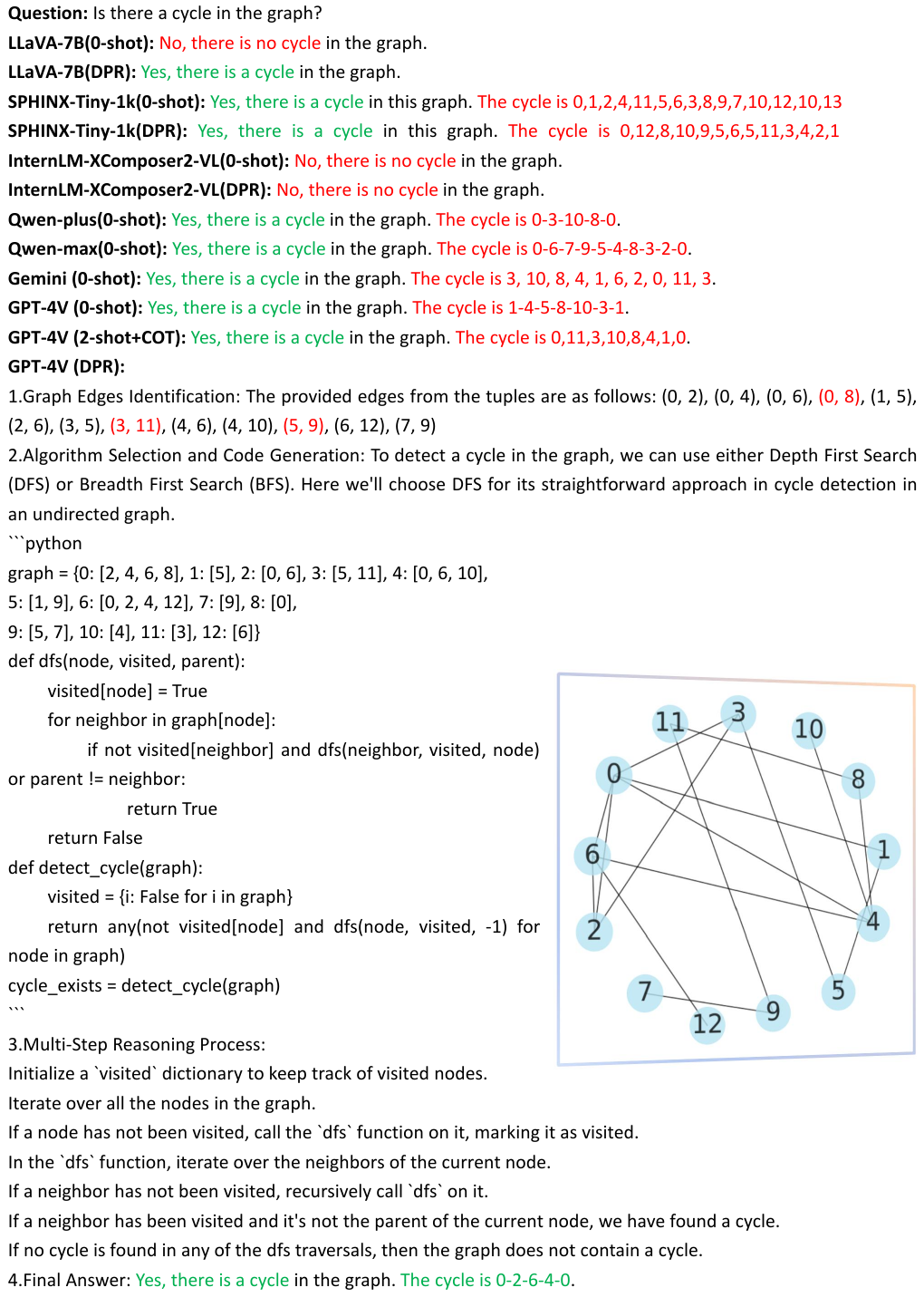}
    \caption{A case (Cycle) illustrates results generated by different models. \textcolor{red}{Red} and \textcolor{green}{green} words are incorrect and correct contents, respectively.}
    \label{fig:cases_study_3}
\end{figure*}
\begin{figure*}[t]
    \centering
    \includegraphics[width=0.85\textwidth]{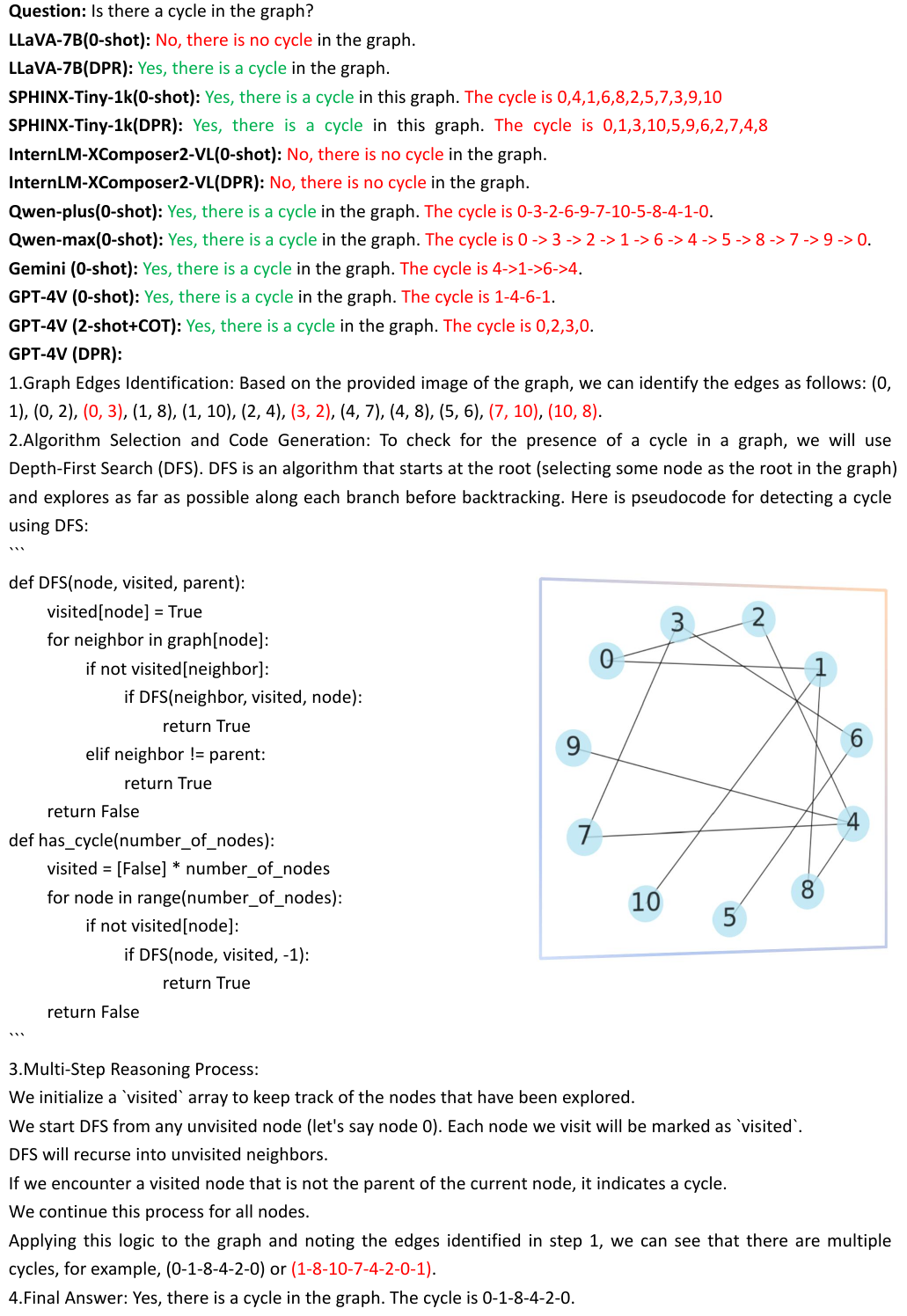}
    \caption{A case (Cycle) illustrates results generated by different models. \textcolor{red}{Red} and \textcolor{green}{green} words are incorrect and correct contents, respectively.}
    \label{fig:cases_study_4}
\end{figure*}
\begin{figure*}[t]
    \centering
    \includegraphics[width=0.85\textwidth]{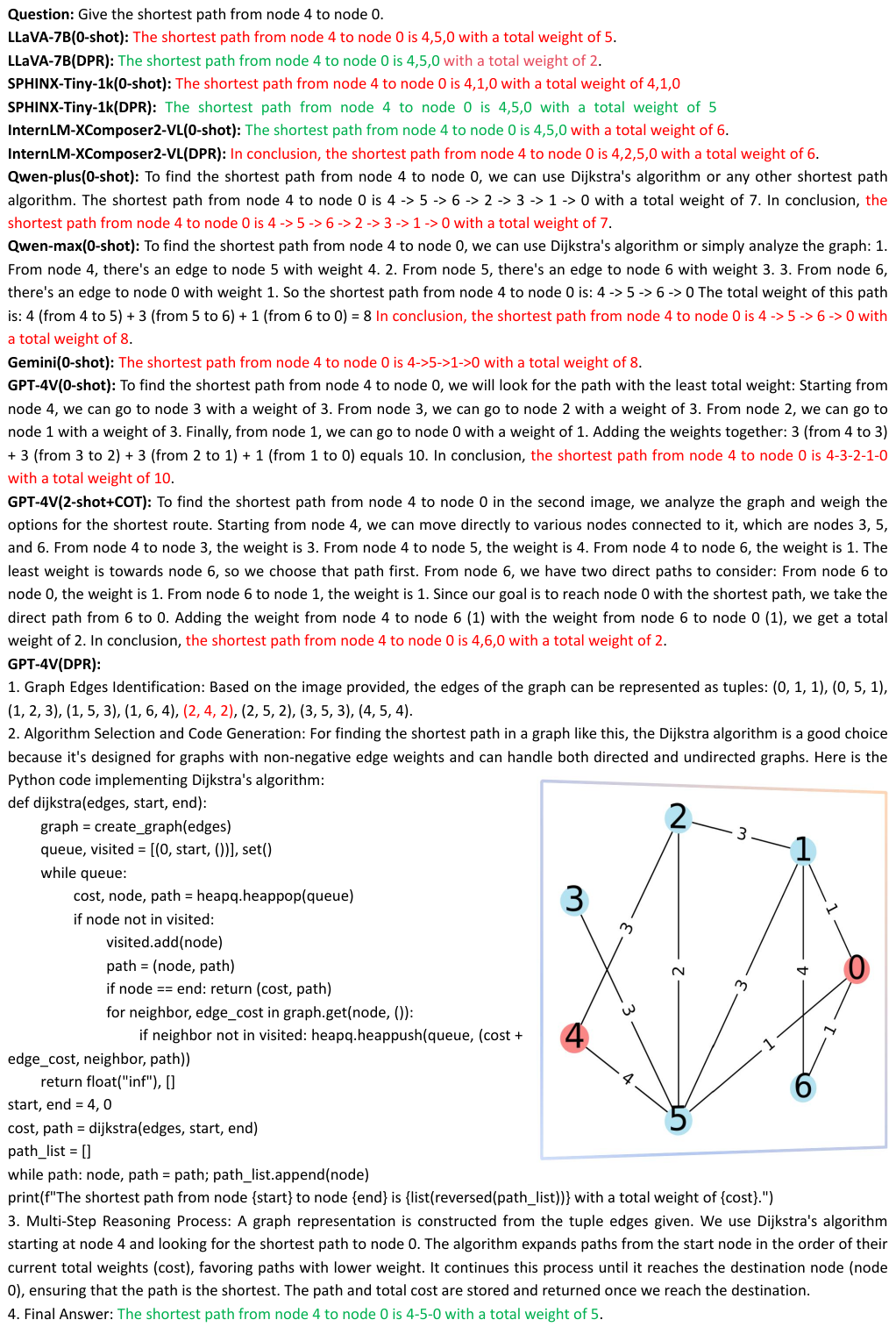}
    \caption{A case (Shortest Path) illustrates results generated by different models. \textcolor{red}{Red} and \textcolor{green}{green} words are incorrect and correct contents, respectively.}
    \label{fig:cases_5}
\end{figure*}

\end{document}